\documentclass[10pt,journal,compsoc]{IEEEtran}
\ifCLASSOPTIONcompsoc
  \usepackage[nocompress]{cite}
\else
  \usepackage{cite}
\fi
\ifCLASSINFOpdf
\else
\fi
\usepackage{amsmath}
\usepackage{graphicx}
\usepackage{bbold}
\usepackage{ragged2e}
\usepackage{multirow}
\newcommand{\bigcell}[2]{\begin{tabular}{@{}#1@{}}#2\end{tabular}}

\newcommand{\Tref}[1]{Table~\ref{#1}}
\newcommand{\Eref}[1]{Eq.~(\ref{#1})}
\newcommand{\Fref}[1]{Fig.~\ref{#1}}
\newcommand{\Sref}[1]{Sec.~\ref{#1}}
\newcommand{\etal}{\textit{et~al.}}

\begin{document}
%
\title{Action-Driven Object Detection \\with Top-Down Visual Attentions}
%
%
%
%

\author{Donggeun~Yoo,~\IEEEmembership{Student Member,~IEEE,}
        Sunggyun~Park,~\IEEEmembership{Student Member,~IEEE,}
		 Kyunghyun~Paeng,~\IEEEmembership{Student Member,~IEEE,}
        Joon-Young~Lee,~\IEEEmembership{Member,~IEEE,}
        and~In~So~Kweon,~\IEEEmembership{Member,~IEEE}
\IEEEcompsocitemizethanks{\IEEEcompsocthanksitem D. Yoo, and I.S. Kweon are with KAIST, South Korea.\protect\\
E-mail: dgyoo@rcv.kaist.ac.kr, iskweon@kaist.ac.kr
\IEEEcompsocthanksitem S. Park and K. Paeng are with Lunit Inc., South Korea.\protect\\
E-mail: sgpark@lunit.io, khpaeng@lunit.io
\IEEEcompsocthanksitem J.-Y. Lee is with Adobe Research, CA, USA.\protect\\
E-mail: jolee@adobe.com}}

\IEEEtitleabstractindextext{
\begin{abstract}
\justifying 
A dominant paradigm for deep learning based object detection relies on a \textit{``bottom-up''} approach using \textit{``passive''} scoring of class agnostic proposals. These approaches are efficient but lack of holistic analysis of scene-level context. In this paper, we present an \textit{``action-driven''} detection mechanism using our \textit{``top-down''} visual attention model. We localize an object by taking sequential actions that the attention model provides. The attention model conditioned with an image region provides required actions to get closer toward a target object. An action at each time step is weak itself but an ensemble of the sequential actions makes a bounding-box accurately converge to a target object boundary. This attention model we call AttentionNet is composed of a convolutional neural network. During our whole detection procedure, we only utilize the actions from a single AttentionNet without any modules for object proposals nor post bounding-box regression. We evaluate our top-down detection mechanism over the PASCAL VOC series and ILSVRC CLS-LOC dataset, and achieve state-of-the-art performances compared to the major bottom-up detection methods. In particular, our detection mechanism shows a strong advantage in elaborate  localization by outperforming Faster R-CNN with a margin of +7.1\% over PASCAL VOC 2007 when we increase the IoU threshold for positive detection to 0.7.
\end{abstract}

}

\maketitle

\IEEEdisplaynontitleabstractindextext

%
\IEEEpeerreviewmaketitle

\IEEEraisesectionheading{\section{Introduction}
\label{sec:introduction}}
\IEEEPARstart{W}{ith} the recent advance \cite{krizhevsky2012imagenet} of deep convolutional neural network (CNN) \cite{lecun1989backpropagation}, CNN based object classification methods in computer vision community have reached human-level performances on the ILSVRC \cite{russakovsky2015imagenet} classification task; 3.57\% \cite{he2016deep} and 3.58\% \cite{szegedy2015rethinking} in top-5 error which are even superior to human showing 5.1\% error \cite{russakovsky2015imagenet}. Thus, current research focus in visual recognition is quickly moving towards richer image understanding problems such as object detection, pixel-level semantic segmentation, image description and question answering in a natural language. Our focus is lying on the object detection problem.

There has been a long line of successful works for object detection \cite{viola2001rapid,lienhart2002extended,vedaldi2009multiple,dollar2009pedestrian,felzenszwalb2010object,bourdev2010detecting,uijlings2013selective} but significant progress in terms of accuracy and efficiency has been achieved by deep learning approaches \cite{szegedy2013deep,sermanet2013overfeat,girshick2014rich,erhan2014scalable,girshick2015fast,ren2015faster,Redmon_2016_CVPR,liu2015ssd} for quite recent years. Among a large literature on object detection with deep learning approaches, one major state-of-the-art family \cite{girshick2014rich,erhan2014scalable,girshick2015fast,ren2015faster} is in Region CNN (R-CNN) pipeline; extracting class agnostic object proposals, applying object classifiers and refining bounding-boxes. The researches \cite{he2015spatial,ouyang2014deepid,simonyan2014very,szegedy2015going} that incorporate R-CNN \cite{girshick2014rich} reported top scores in ILSVRC'14 and Faster R-CNN \cite{ren2015faster} won at ILSVRC'15. However, even the most accurate and efficient R-CNN pipeline embeds a limitation of not reflecting the important visual contexts outside a proposal, caused by the passive scoring of the proposal with classifiers. To avoid such limitation, there are alternative top-down approaches~\cite{gonzalez2015active,yoo2015attentionnet,caicedo2015active,Mathe_2016_CVPR} which actively explore the location of a target object by taking surrounding context into account. However, the top-down approach for deep learning based object detection has not been much investigated yet.

In this paper, we propose an action-driven method for top-down object detection. We cast an object detection problem as a sequential action problem. We introduce a visual attention model named AttentionNet which acts as an agent determining what action should be taken in the next step. This model takes an image region as input and provides the optimal actions for getting closer toward a target object. In our detection mechanism, this attention model is fully utilized from beginning to the end of object detection pipeline. Starting from a whole image or a large region, our detection mechanism actively explores the location of a target object and finishes by drawing an accurate bounding-box. The background context surrounding a target object is also taken into consideration since the searching scope in its early stage is broad enough.

\begin{figure*}[t]
\begin{center}
\includegraphics[width=1\linewidth]{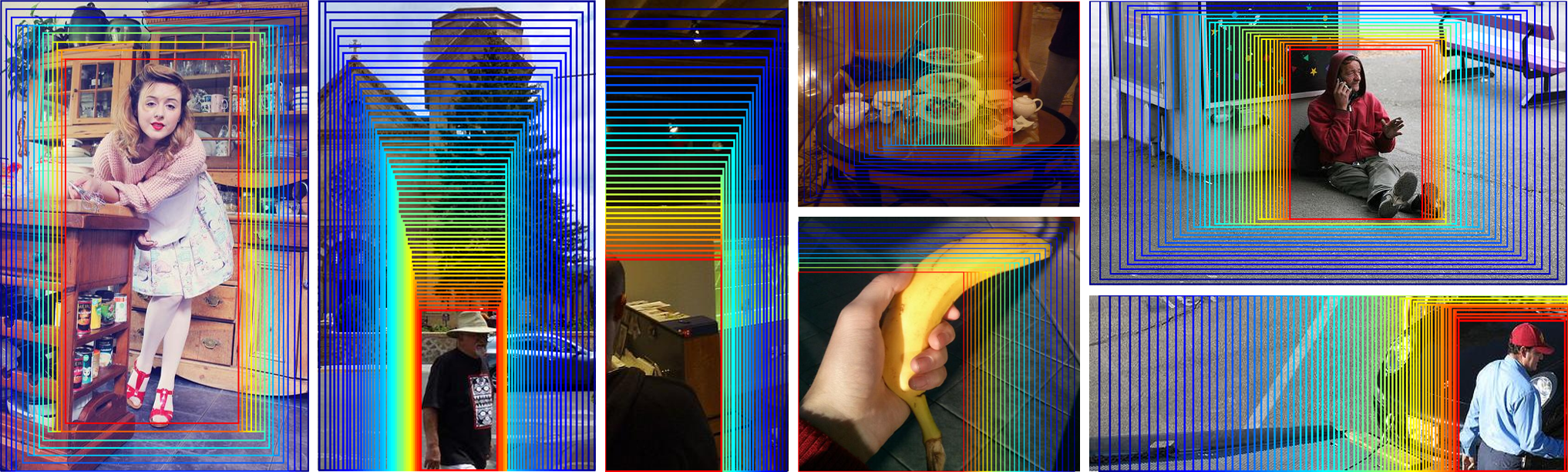}
\end{center}
\caption{Real detection examples of our detection mechanism. Starting from an image boundary (dark blue bounding-box), our detection mechanism recursively narrows the bounding-box down to a final human location (red bounding-box). For visual effects, we set the size of each movement small (5 pixels given a 224$\times$224 size input) for all the examples.}
\label{fig:teaser}
\end{figure*}

The core of our detection mechanism lies on an idea of taking an action sequence by order of the attention model. The attention model tells us a pair of actions, which should be taken at the top-left and bottom-right corner of an input image, to get closer to a target object. For instance, the action could be ``go down'' or ``go left'' at each corner respectively. We then simply take the actions by cropping the input image until the image boundary converges to a target object. Even if each action is inaccurate, taking a set of multiple actions results in an accurate boundary of a target object, such as an ensemble method combines many weak learners to produce a strong learner. \Fref{fig:teaser} shows real examples of our detection mechanism. Starting from an entire image, the bounding-box moves sequentially then converges to a target object boundary.

Our detection mechanism is radically distinct from the state-of-the-art R-CNN based methods. These methods depend on the bottom-up object proposals and score them with classifiers, while we follow a top-down search strategy. The bottom-up object proposals are based on the characteristic of a local scene so called ``objectness''. Proposals of \cite{gu2009recognition,uijlings2013selective,cheng2014bing,zitnick2014edge} are driven from low-level features and those of \cite{erhan2014scalable,ren2015faster} are from trainable mid-level features. In contrast, our top-down mechanism is controlled by high-level sub-tasks, i.e. detection boxes are driven from a sequence of actions. The bottom-up approaches are inherently faster than our top-down approach since they are feed-forward while ours is recurrent. However, the top-down approach has its strong property coming from high-level reasoning, such that the context surrounding a target object could be reflected to the action sequence. Thus, this top-down approach can be a complementary way toward the next direction of object detection.

Our detection mechanism with a single attention model does everything necessary for a detection pipeline but yields state-of-the-art performance. With a single attention model, we 1) detect initial regions where a single instance is included, 2) detect objects by taking sequential actions from each initial region, and 3) finally refine the localizations by taking an extra action sequence. Therefore, we do not incorporate any separate modules for object proposals nor post bounding-box regression.

A preliminary version of this work was published in \cite{yoo2015attentionnet}, which can detect only a single object class. Since then we have generalized the detection mechanism to handle multiple object classes with a shared attention model. Also, we have given important modifications to determining the scaling factors used for multi-scale training and inference.

\subsection{Contributions} In summary, our contributions are three-fold.
\begin{enumerate}
\item We suggest an action-driven object detection mechanism, which actively search exact object locations by taking action sequence produced by an attention model.
\item The detection mechanism does not requires any separated modules for object proposal nor post bounding-box regression. Taking sequential actions covers all these.
\item This is a top-down detection mechanism which first achieves the state-of-the-art accuracy compared to the recent bottom-up methods.
\end{enumerate}

\section{Related Works}
\label{sec:related}
There has been a large literature on object detection for the last few decades. Object models are learned from low-level features \cite{viola2001rapid,dollar2009pedestrian} or mid-level part based features \cite{felzenszwalb2010object,bourdev2010detecting}, and the models evaluate image regions in a sliding window fashion. Since then, a raise of object proposal methods \cite{gu2009recognition,uijlings2013selective,cheng2014bing,zitnick2014edge}, which generate thousands of potential bounding-boxes, substantially improves detection efficiency compared to the sliding window search. We refer readers to \cite{hosang2016makes} for an in-depth study on various object proposal methods.

In recent researches of object detection, a significant progress in both of accuracy and efficiency has been achieved by a powerful combination of all three; high-quality object proposals \cite{uijlings2013selective}, a deep network to represent the proposals \cite{krizhevsky2012imagenet}, and big data for training the network \cite{deng2009imagenet}. This framework called R-CNN was proposed by Girshick~\etal~\cite{girshick2014rich} and has become a dominant paradigm for object detection. From here we limit our review to the deep learning approaches. 

The successful performance of the R-CNN pipeline triggered engineering challenges to make it run in real-time. An attempt to train a feed-forward network for object proposal \cite{erhan2014scalable} speeds up the proposal step but requires per-region classification which is far from a real-time speed. In contrast, per-region pooling over a shared convolution feature map \cite{he2015spatial,girshick2015fast} substantially boosts the speed for classifying the proposals but extracting the proposal \cite{uijlings2013selective} is a bottleneck. Ren~\etal~\cite{ren2015faster} design a region proposal network, and make this network and a classification network share the convolution feature maps. This system called Faster R-CNN finally performs in near real-time with an improved accuracy.

These bottom-up approaches are inherently feed-forward and fast but lack of holistic analysis of scene-level context. Our top-down approach is relatively slow due to the recurrent actions but can see the larger context while taking actions. There has been three recent works \cite{gonzalez2015active,caicedo2015active,Mathe_2016_CVPR} similar to ours in terms of adopting the action-driven top-down approach. Gonzalez-Garcia~\etal~\cite{gonzalez2015active} propose an active search strategy which depends on the spatial context and the region scores in the previous state. Caicedo and Lazebnik~\cite{caicedo2015active} and Mathe~\etal~\cite{Mathe_2016_CVPR} also present action-driven detection methods in which an agent determining actions is trained by reinforcement Q-learning \cite{sutton1998reinforcement}. These three works successfully apply the top-down approach to the detection problem, however, the performances are still far from state-of-the-art competitors, and the detector of~\cite{caicedo2015active} is class-specific. In this paper, we extend our previous class-specific model \cite{yoo2015attentionnet} to handle multiple classes and achieve state-of-the-art performances.

We introduce another side of detection paradigm where a detection problem is framed as a regression problem. A feed-forward network directly estimates bounding-boxes. Szegedy~\etal~\cite{szegedy2013deep} trains a deep network which maps an image to a rectangular mask of an object. Sermanet~\etal~\cite{sermanet2013overfeat} also employ a similar approach but their network directly estimates bounding-box coordinates. These models produce a single bounding-box so should be evaluated on sliding windows to detect multiple instances. Quite recently, Redmon~\etal~\cite{Redmon_2016_CVPR} and Liu~\etal~\cite{liu2015ssd} develop a regression model that produces multiple bounding-boxes and their class probabilities. They estimate a bounding-box for each grid cell of a convolution feature map, so all the outputs are obtained in a single feed-forward path. All these detection-by-regression approaches are also related to our work in that they do not rely on object proposals and actively produce bounding-boxes. However, ours is distinct from these methods in that our regression proceeds sequentially with high-level reasoning.

We use an attention model as an agent, so a long line of the attention models is also related to ours. For recent years, incorporating the visual attention idea into a deep network \cite{ba2014multiple,mnih2014recurrent,gregor2015draw,jaderberg2015spatial} has been proposed to select regions that need more attention for better visual recognition such as classification \cite{jaderberg2015spatial,xiao2015application,cao2015look} and caption generation \cite{xu2015show,yao2015describing}. Our use of the attention model differs from them in its aim and supervision. We incorporate the attention model in order to determine the optimal action to get closer to an object whereas their models aim at more focused representation to help their target recognition tasks. Also, we can train our model in a supervised fashion with optimal action labels determined from ground-truth bounding-boxes. In contrast, attention of these methods cannot be directly supervised since their target labels do not include locations of significant objects. For this reason, they often employ the reinforcement Q-learning, or design a differentiable model that could be optimized with a back-propagation in a weakly supervised fashion.

This paper is organized as follows. We first introduce our class-specific detection mechanism in \Sref{sec:anet:method} and evaluate the method by a human detection task in \Sref{sec:anet:exp}. We then generalize the detection mechanism to multiple object classes in \Sref{sec:manet:method} and also evaluate that in \Sref{sec:manet:exp}. We finally conclude this study in \Sref{sec:conclusion}.

\section{Detection Mechanism}
\label{sec:anet:method}
We introduce our detection mechanism under an assumption that an input image includes a single object instance only. Extension of the mechanism to multiple instances will be described in \Sref{sec:anet:method:glance}.
\begin{figure*}[t]
\begin{center}
\includegraphics[width=0.88\linewidth]{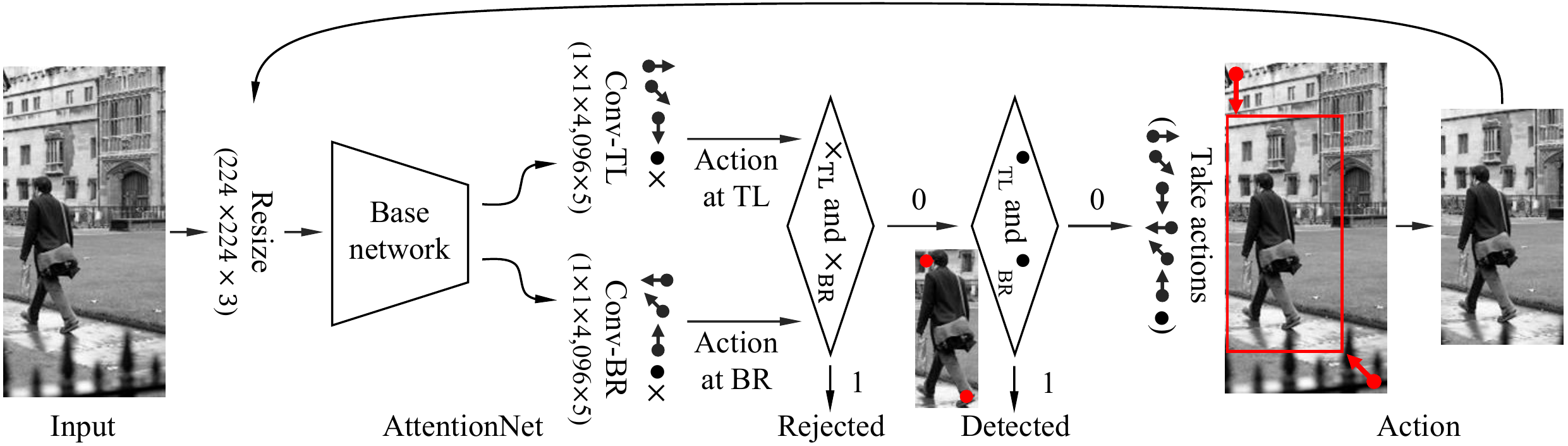}
\end{center}
\caption{Action-driven detection mechanism. The attention model tells us a pair of actions which should be taken at the top-left corner (TL) and the bottom-right corner (BR) of an input image. The action set for TL is defined as follows; go right $\rightarrow$, go right-down $\searrow$, go down $\downarrow$, stop $\bullet$ and reject $\times$. The action set for BR is also defined in this way but with opposite directions. If the attention model produces the action ``reject $\times$'' in both corners, we reject the input. If not, we apply the actions to the input and feed it to the model again until it meets the action ``stop $\bullet$'' in both corners.}
\label{fig:anet:method:pipeline}
\end{figure*}

\Fref{fig:anet:method:pipeline} shows how we frame an object detection problem as a sequential action problem. We first warp an input to a fixed size image and feed it to the attention model named AttentionNet. The attention model then tells us a pair of actions required for the input to get closer to the target object. The actions will be applied to the top-left corner (TL) and the bottom-right corner (BR) of the input image respectively. We define a high-level action set for TL as follows; go right $\rightarrow$, go right-down $\searrow$, go down $\downarrow$, stop $\bullet$ and reject $\times$. We also define the action set for BR in this way but the directions of the movement actions are opposite to those of TL.

The attention model is indicating $\downarrow_{\text{TL}}$ and $\nwarrow_{\text{BR}}$ in \Fref{fig:anet:method:pipeline}. We then apply the actions at both corner in the way of cropping the input. The amount of movement $l$ is constant. The cropped image is fed to the attention model again until the image meets one of the two terminal conditions; $\times$ at both corners, or $\bullet$ at both corners. An image given $\times$ at both corners is regarded as a background while an image given $\bullet$ at both corners is a detection result. The detected image boundary is back-projected to a bounding-box in the original input image domain. Given a stopped (detected) bounding-box $b$ and its corresponding output activations $\mathbf{y}_{\text{TL}}, \mathbf{y}_{\text{BR}}\in \Re^5$ before a softmax normalization, the detection score $s^b$ is discriminatively defined as
\begin{equation}
\begin{split}
s^b&=\frac{1}{2}\left(s^b_{\text{TL}}+s^b_{\text{BR}}\right),\quad \text{s.t.}\\
s^b_{\text{TL}}&=y^{\bullet}_{\text{TL}}-(y^{\rightarrow}_{\text{TL}}+y^{\searrow}_{\text{TL}}+y^{\downarrow}_{\text{TL}}+y^{\times}_{\text{TL}}),\\
s^b_{\text{BR}}&=y^{\bullet}_{\text{BR}}-(y^{\leftarrow}_{\text{BR}}+y^{\nwarrow}_{\text{BR}}+y^{\uparrow}_{\text{BR}}+y^{\times}_{\text{BR}}).
\label{eq:anet:method:score}
\end{split}
\end{equation}

Compared with the R-CNN framework which depends on object proposals, our detection starts from a large area and actively reaches at a terminal point with \textit{stop signals}. In early stage of this procedure, we can take the \textit{large context} surrounding an object into consideration. Such a large context is an important cue for identifying the class of the object. This benefit will be highlighted with an experiment in \Sref{sec:anet:method:toy} again. 

Compared with the previous detection-by-regression approaches \cite{szegedy2013deep,sermanet2013overfeat,Redmon_2016_CVPR,liu2015ssd}, we solve the regression problem by iterative classifications of high-level actions. Even if the actions in early stage could be inaccurate, subsequent actions become stronger as the searching scope is gradually narrowed down to an object.

\subsection{Attention Model}
\label{sec:anet:model}
Our detection mechanism requires an agent which determines the optimal actions to be applied to both corner of an input. The agent can be a regression model that tells us a location coordinate but the regression is a more difficult task for a network compared to a classification task that classifies quantized directions. Thus we choose a classification model, which is trainable with a softmax loss. The end of the model is composed of two classification layers for both corners, and each layer classifies the five actions including the three movement actions ($\rightarrow,\searrow,\downarrow$ for TL) and the two termination actions ($\bullet,\times$). The five fully connected filters of 1$\times$1$\times$4,096 size determine the five action scores for each corner. We can choose a base network for this model from any popular convolutional network architectures. The illustration of this model is shown in \Fref{fig:anet:method:pipeline}. 

\subsection{Training}
\label{sec:anet:method:training}
The required actions are determinant to the location of a target object. Always we can determine the optimal action for an arbitrary region to get closer to the ground-truth bounding-box. Selecting the optimal action at each time step does not depend on the previous action sequence. Thus, we can train our attention model regardless of that.

Caicedo and Lazebnik~\cite{caicedo2015active} also present an action-driven detection mechanism with an agent. Their action set are differently defined such as horizontal moves, vertical moves, scale changes and aspect ratio changes. Given this action set, they adopt the reinforcement Q-learning~\cite{sutton1998reinforcement} with an IoU (Intersection over Union) based reward. Despite their interesting application of reinforcement learning for object detection, the reinforcement method inherently has a high variance in the gradient of the expected reward so it is difficult to accurately find a Q value which is approximated by a deep network with a limited size of training set. In contrast, since our actions are designed to be optimally chosen to increase the IoU at any states, we can train our attention model with the softmax loss.

To make the attention model operate in the scenario we devise, it is important to process original training images to a suitable form. During the inference stage, the number of possible action pairs is 17 ($=$4$\times$4$+$1) such as $\left\{\rightarrow,\searrow,\downarrow,\bullet\right\}_{\text{TL}}\times\left\{\leftarrow,\nwarrow,\uparrow,\bullet\right\}_{\text{BR}}$ for positive regions and $\left\{\times_{\text{TL}},\times_{\text{BR}}\right\}$ for negative regions. To evenly cover these 17 cases in training, we augment the original training images into a reformed region set. 

\begin{figure}[t]
\begin{center}
\small
\includegraphics[width=1\linewidth]{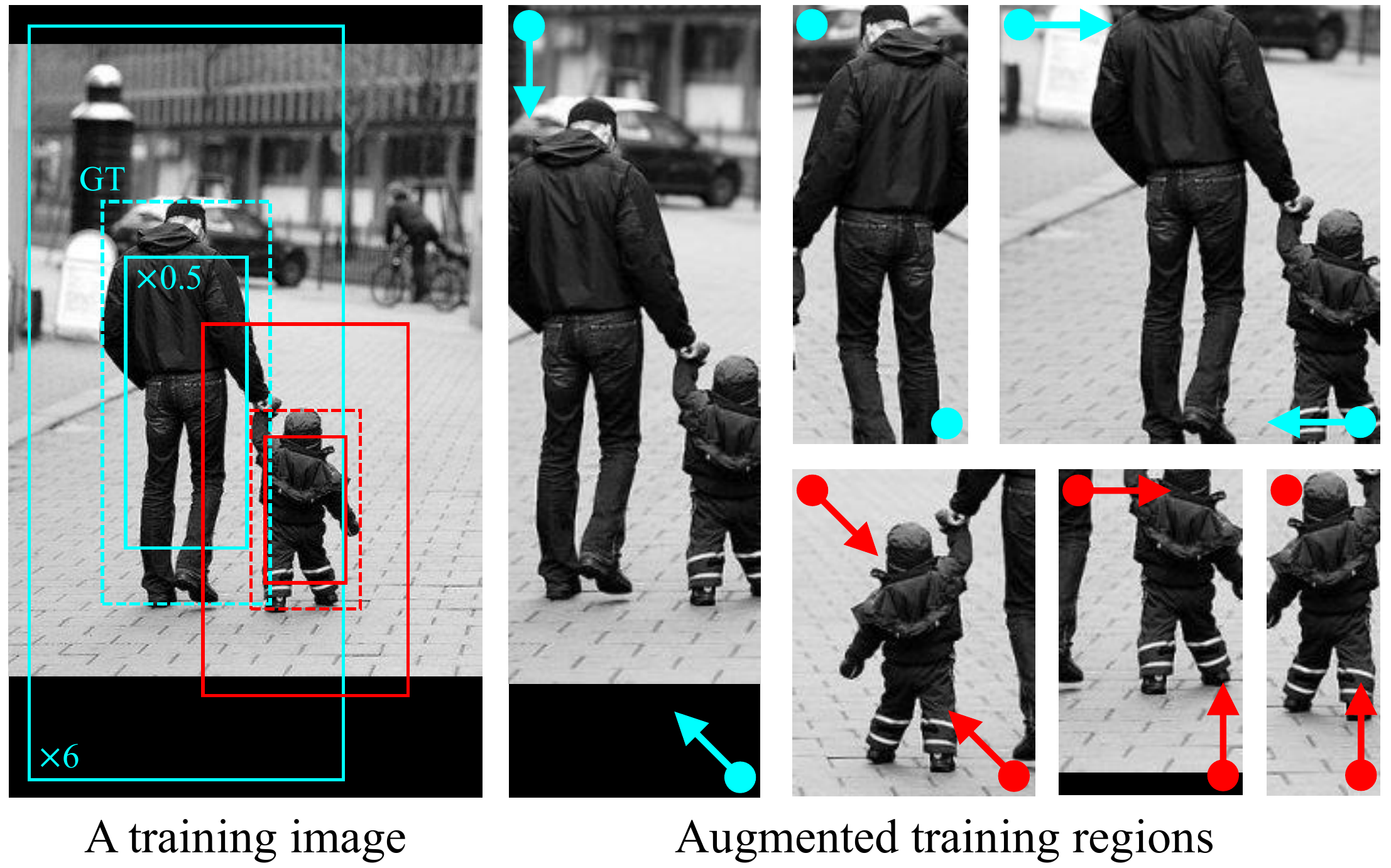}
\end{center}
\caption{Examples of generating training regions to learn the attention model. A dashed rectangle is a ground-truth. The box inside a ground-truth is an inner bound which is 2 times smaller than the ground-truth while the outer bound is 6 times larger than that. Any region between these two bounds is randomly sampled with a random horizontal flip. A ground-truth action label at each corner is assigned automatically. The area beyond the image boundary is filled with zeros.}
\label{fig:anet:method:training}
\end{figure}

\Fref{fig:anet:method:training} shows examples how we process an original training image to multiple augmented regions. We randomly sample any regions between the inner and outer bound. The inner bound is 2 times smaller than the ground-truth (a dashed box) whereas the outer bound is 6 times larger. The area out of an image boundary is filled with zeros. It is important to make the outer bound sufficiently large to take the \textit{large context} into account during training. Each region is horizontally flipped with a probability of 0.5. We then assign a pair of ground-truth action labels at both corners that is determined from a ground-truth bounding-box. We also randomly generate negative regions which are not overlapped with ground-truth bounding-boxes.

Some regions probably include multiple instances as in the most top-right example in \Fref{fig:anet:method:training}. In this case, we simply assign action labels for the biggest instance. These regions are also essential for training. Let us consider a multiple instance detection scenario. If an attention model is trained without these regions, a final detection result from a large initial region probably includes the multiple instances at ones. To make our mechanism always narrow the box down to the biggest instances within the visible area, we must make the outer bound sufficiently large ($\times$6).

When we compose a mini-batch for training, we select positive and negative regions in an equal portion. In a batch, each of the 16(=4$\times$4) cases for positive regions occupies a portion of 1/(2$\times$16), and the negative regions occupy the remaining portion of 1/2. The loss for training is an average of the two log-softmax losses computed independently in TL and BR
\begin{multline}
\ell=\frac{1}{2}\cdot\ell_{\text{softmax}}\left(\mathbf{y}_{\text{TL}}, t_{\text{TL}}\right)+\frac{1}{2}\cdot\ell_{\text{softmax}}\left(\mathbf{y}_{\text{BR}}, t_{\text{BR}}\right)\\
\text{s.t.}\quad\ell_{\text{softmax}}\left(\mathbf{y}, t\right)=-y_t+\log\sum_{i}e^{y_i}
\label{eq:multi-loss}
\end{multline}
where $\mathbf{y}$ is a 5-dimensional action score vector and $t$ is a ground-truth action label index.

\subsection{Top-down VS. Bottom-up}
\label{sec:anet:method:toy}
Before we make our detection mechanism detect multiple instances, we verify the effectiveness of our top-down approach, against to the bottom-up approach relying on region proposals~\cite{uijlings2013selective}. As studied by \cite{agrawal2014analyzing}, strong mid-level activations in a deep network come from object parts that is distinctive to other object classes. Since R-CNN based detection depends on the score computed with the activations inside each proposal, the results often focuses on discriminative object \textit{parts} (e.g. face) rather than an entire object (e.g. entire human body).

To analyze this issue, we design an experiment of \textit{single}-human detection over PASCAL VOC 2007~\cite{Everingham10}. We train an AlexNet~\cite{krizhevsky2012imagenet} based R-CNN human detector with the code provided by the authors~\cite{girshick2014rich}. Also, we train an AlexNet based attention model with the same training data. From VOC 2007 \texttt{test} set, we choose images that contain a single human instance to make a sub test set. To highlight the only difference that comes from top-down and bottom-up approaches, we just choose the \mbox{top-1} region as a detection result for R-CNN. Our detection mechanism begins with the entire image boundary and detects an instance. We measure average precisions (AP) with a standard IoU threshold of 0.5 for positive detection.

\begin{figure}[t]
\begin{center}
\includegraphics[width=0.98\linewidth]{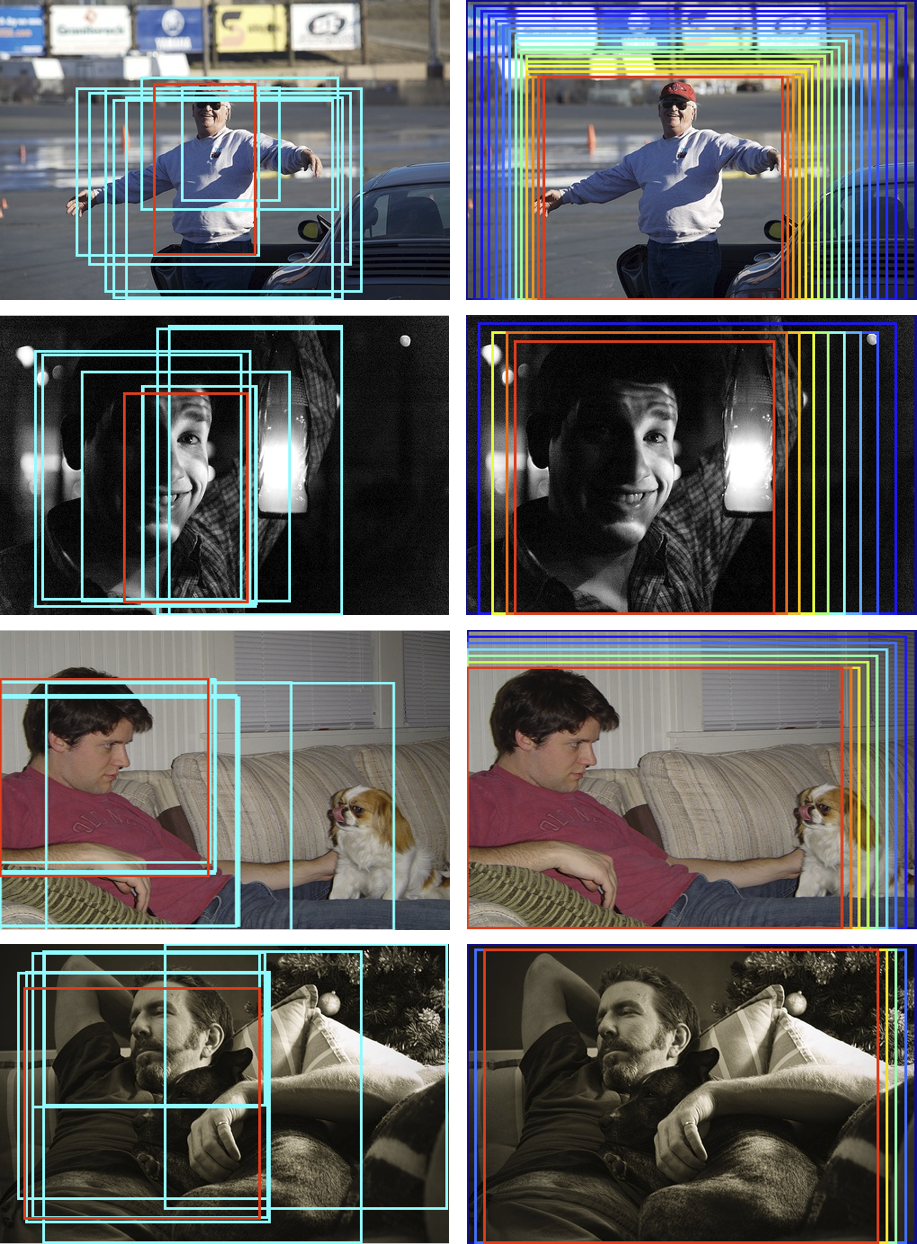}
\end{center}
\caption{Qualitative comparison of the bottom-up and top-down approach with a \textit{single}-human detection task. The left column is the bottom-up R-CNN based method and the right column is our top-down method. In the left column, a red bounding-box is the top-1 region and the cyan boxes are top-10 regions.}
\label{fig:anet:method:toy}
\end{figure}

\Tref{tab:anet:method:toy} shows the results. The bottom-up approach shows 79.4\% while our top-down approach shows 89.5\%. The bottom-up approach shows much lower detection performance due to the weak correlation between a classification score of a proposal and the \textit{entire} human body. As shown in \Fref{fig:anet:method:toy}, the maximally scored object proposal of R-CNN is prone to focus on the discriminative faces rather than the entire human bodies. In contrast, our detection mechanism reaches a terminal point starting from a boundary out of a target object.
\begin{table}[t]
\begin{center}
\begin{tabular}{|l|l|c|}\hline
Method&Approach&AP (\%)\\\hline\hline
Top-1 result from R-CNN\cite{girshick2014rich}&Bottom-up&79.4\\
AttentionNet&Top-down&\textbf{89.5}\\\hline
\end{tabular}
\end{center}
\caption{\textit{Single}-human detection performances of the bottom-up and top-down approaches on a subset of PASCAL VOC 2007 \texttt{test} set, in which each image contains a single human instance.}
\label{tab:anet:method:toy}
\end{table}

\subsection{Initial Glance}
\label{sec:anet:method:glance}
Our attention model provides actions toward a single instance of a visible region. In this section, we introduce an efficient method to extend our detection mechanism to a practical scenario in which an image includes multiple instances. Our solution is to initialize a large box for each instance. We call this initial glance. To this end, we also utilize our attention model, therefore, a separated model is unnecessary. We then can detect an instance from each initial glance, and merge the results into a reduced number of bounding-boxes followed by a final refinement procedure for which we reuse the attention model again.

\begin{figure}[t]
\begin{center}
\footnotesize
\includegraphics[width=0.98\linewidth]{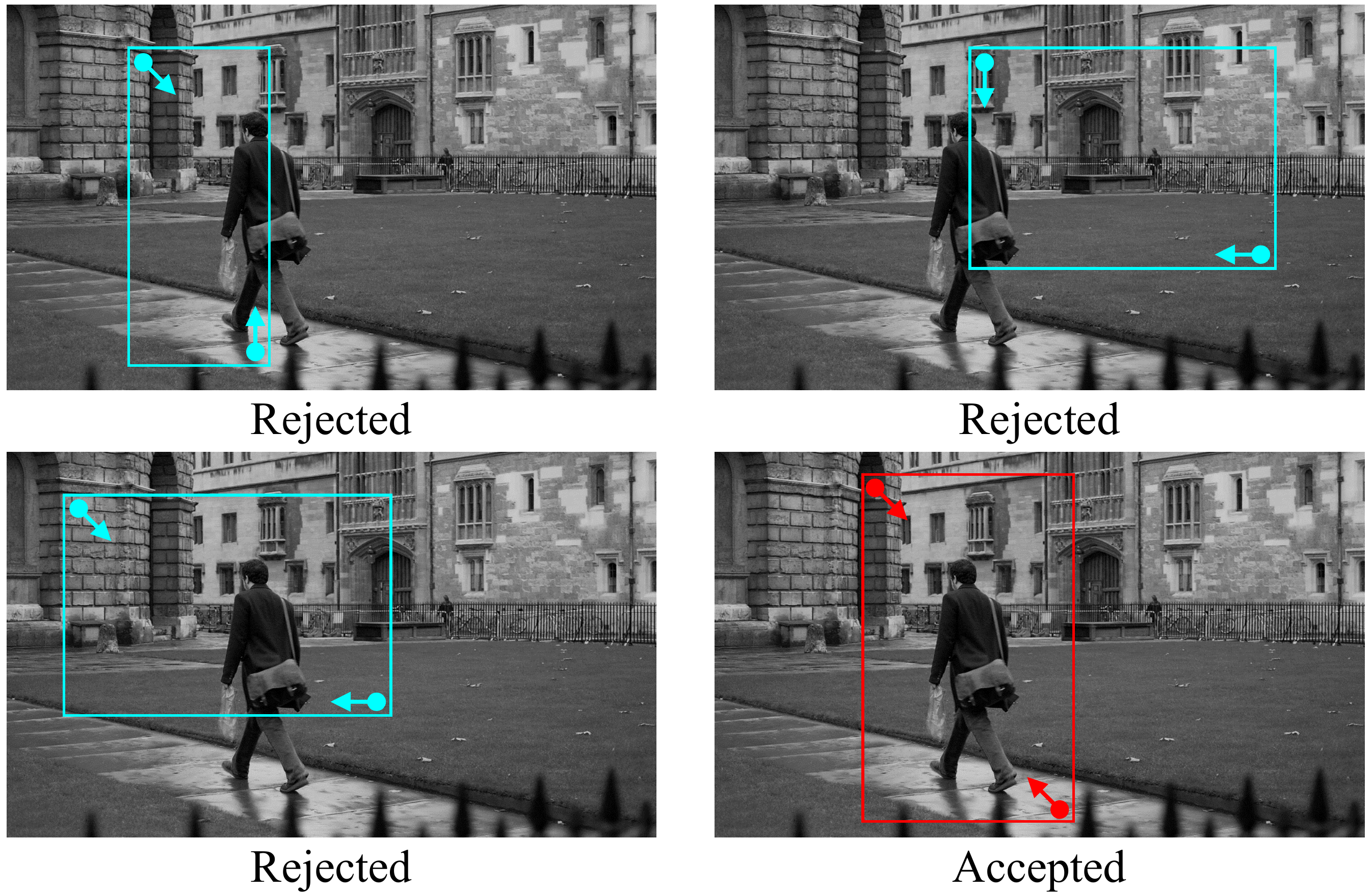}
\end{center}
\caption{A condition for a window to be an initial glance. Among muti-scale and muti-aspect ratio windows, we choose only regions that predicted as ($\searrow$,$\nwarrow$) at each corner as initial glances to make sure that the entire object instance is included.}
\label{fig:anet:method:glance}
\end{figure}

A required condition for a region to be an initial glance is that the region should contain an entire instance with sufficient surrounding contexts. Let us assume we have an arbitrary region in an image and we feed this region to the attention model. Among 17 possible action combinations, the action prediction of ($\searrow_{\text{TL}},\nwarrow_{\text{BR}}$) guarantees that the region includes the entire body of an instance with enough margins. In the other predictions, it is possible for a region to be truncating an instance or be a background. Some examples are shown in \Fref{fig:anet:method:glance}. 

\begin{figure}[t]
\begin{center}
\includegraphics[width=1\linewidth]{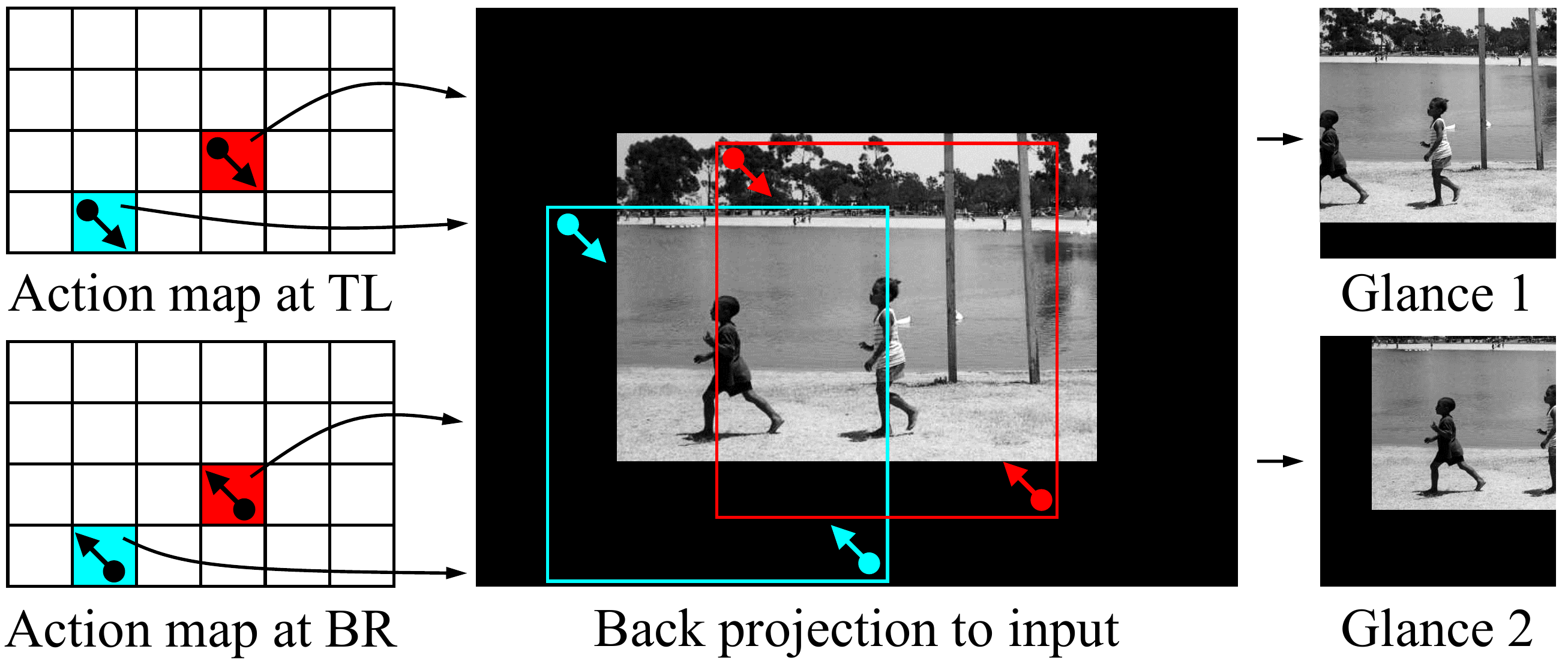}
\end{center}
\caption{Efficient initial glance mining from action maps. A large image is fed to the fully convolutional attention model to obtain an action map for each corner. Regions satisfying the condition of ($\searrow_{\text{TL}},\nwarrow_{\text{BR}}$) becomes initial glances. The glances are given to the attention model again to detect each instance as illustrated in \Fref{fig:anet:method:pipeline}.}
\label{fig:anet:method:glance-mining}
\end{figure}

To boost speed and recall of the initial glance mining, we follows the fully convolutional technique presented by \cite{tompson2014joint,yoo2015multi,long2015fully}. An input of a convolutional network is not limited to a fixed size since a fully connected layer could be replaced with a convolution layer containing 1$\times$1 size filters. We feed $K$ multi-scale multi-aspect ratio images to our fully convolutional attention model and obtain $K$ action maps for each corner. For instance, if our base network is \mbox{VGG-16}~\cite{simonyan2014very}, a pixel and its neighbor in an action map correspond to 224$\times$224 size regions with a stride of 32 in an input image. Given these action maps, we choose regions that satisfy ($\searrow_{\text{TL}},\nwarrow_{\text{BR}}$) condition as initial glances. An example is shown in \Fref{fig:anet:method:glance-mining}. The initial glances are then given to the attention model again to detect each instance as the mechanism described in \Fref{fig:anet:method:pipeline}. For some instances located on side of an image, each side of an image dilates with a 112(=224/2) size margin filled with zeros before being fed to the attention model.

Object instances are diverse in aspect ratio and scale. Thus the scales and aspect ratios for an input image are important to successfully mine initial glances in an inference stage. Thus, we introduce a data-driven approach to determine $K$ scales and aspect ratios. Let us assume the minimum input size for an attention model is $(224,224)$. Also, we have an input image of $(w,h)$ size and its ground-truth bounding-box $b$ of $(w_{b},h_{b})$ size. If we rescale this input to $\left(w\cdot\frac{224}{w_{b}},h\cdot\frac{224}{h_{b}}\right)$ size, each pixel in an action map corresponds to a region which has a size equal to the ground-truth bounding-box in the original image domain. To make some regions satisfy ($\searrow_{\text{TL}},\nwarrow_{\text{BR}}$) condition with an enough margin, we can define a margin factor $\alpha>1$ with which we can rescale the input to $\left(w\cdot\frac{224}{\alpha\cdot w_{b}},h\cdot\frac{224}{\alpha\cdot h_{b}}\right)$ size. We define a scaling factor multiplied to $(w,h)$
\begin{equation}
\mathbf{s}_b=\left[s_b^w,s_b^h\right]=\left[\frac{224}{\alpha\cdot w_{b}},\frac{224}{\alpha\cdot h_{b}}\right].
\end{equation}
Our objective here is to determine $K$ representative scaling factors $\left\{\mathbf{s}_k\;|\;k=1\cdots K\right\}$, which will be used for the inference stage, to maximize the chance for mining initial glances.

Given a training image set $I^{tr}$ and their ground-truth bounding-boxes which are $\{w_b, h_b\;|\;b\in I^{tr}\}$-size, we compute scaling factor samples $\left\{\mathbf{s}_b\;|\;b\in I^{tr}\right\}$ for all the bounding-boxes. $K$ representative scaling factors are then estimated by grouping the samples. We run k-means clustering algorithm over the samples $\left\{\mathbf{s}_b\;|\;b\in I^{tr}\right\}$ in a log-scale space and obtain $K$ centroids $\left\{\mathbf{s}_k\;|\;k=1\cdots K\right\}$. In an inference stage, we can rescale a test image to multiple sizes $\left\{w\cdot s_k^w,h\cdot s_k^h\;|\;k=1\cdots K\right\}$ and feed them to the attention model for the initial glance mining. If a rescaled image size is smaller than 224, we pad zeros with equal margins in both sides.

\subsection{Initial Detection and Refinement}
\begin{figure*}[ht]
\begin{center}
\small
\begin{tabular}{ccccc}
\setlength{\tabcolsep}{1.7pt}
\includegraphics[width=0.17\linewidth]{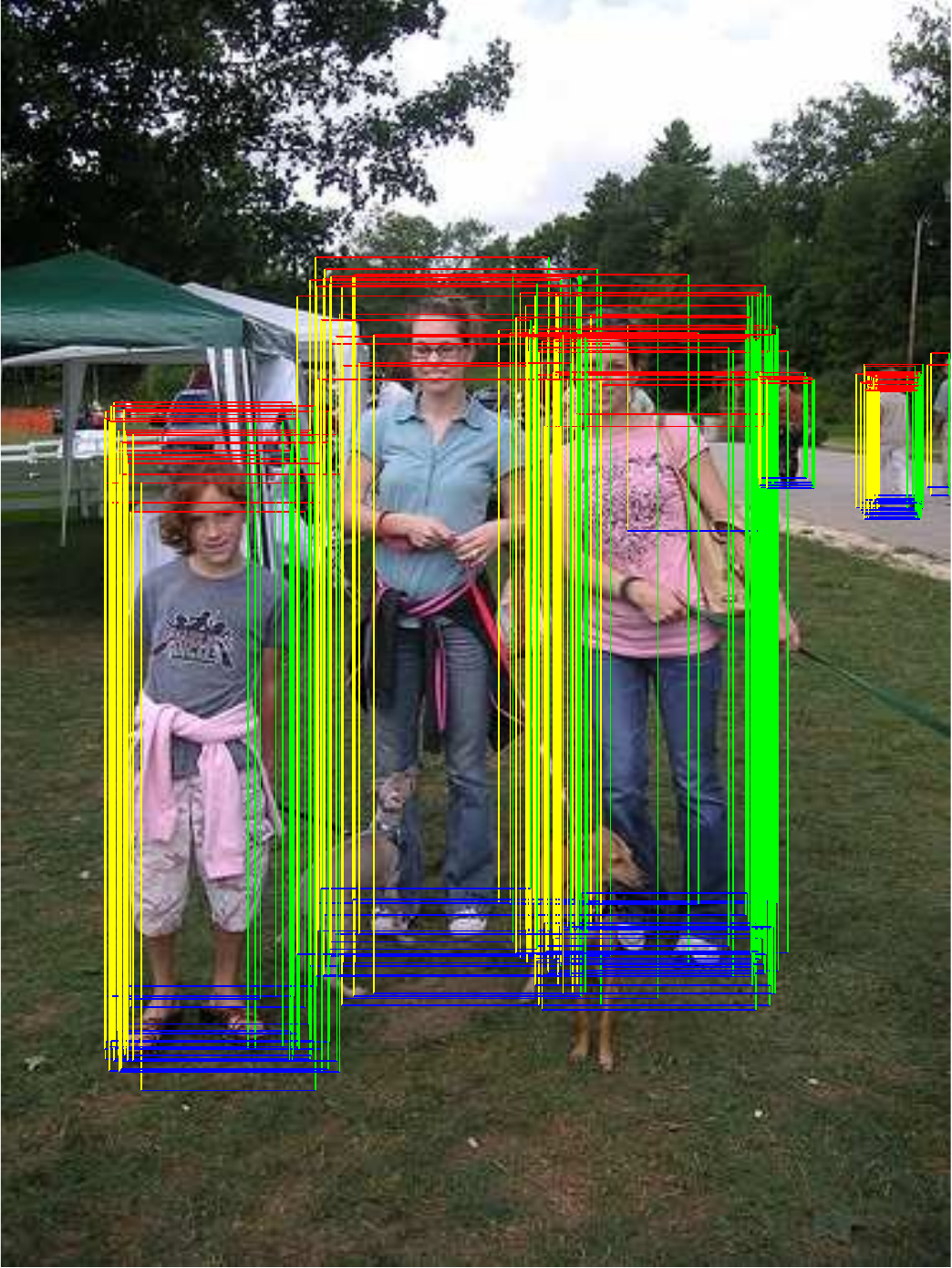}&
\includegraphics[width=0.17\linewidth]{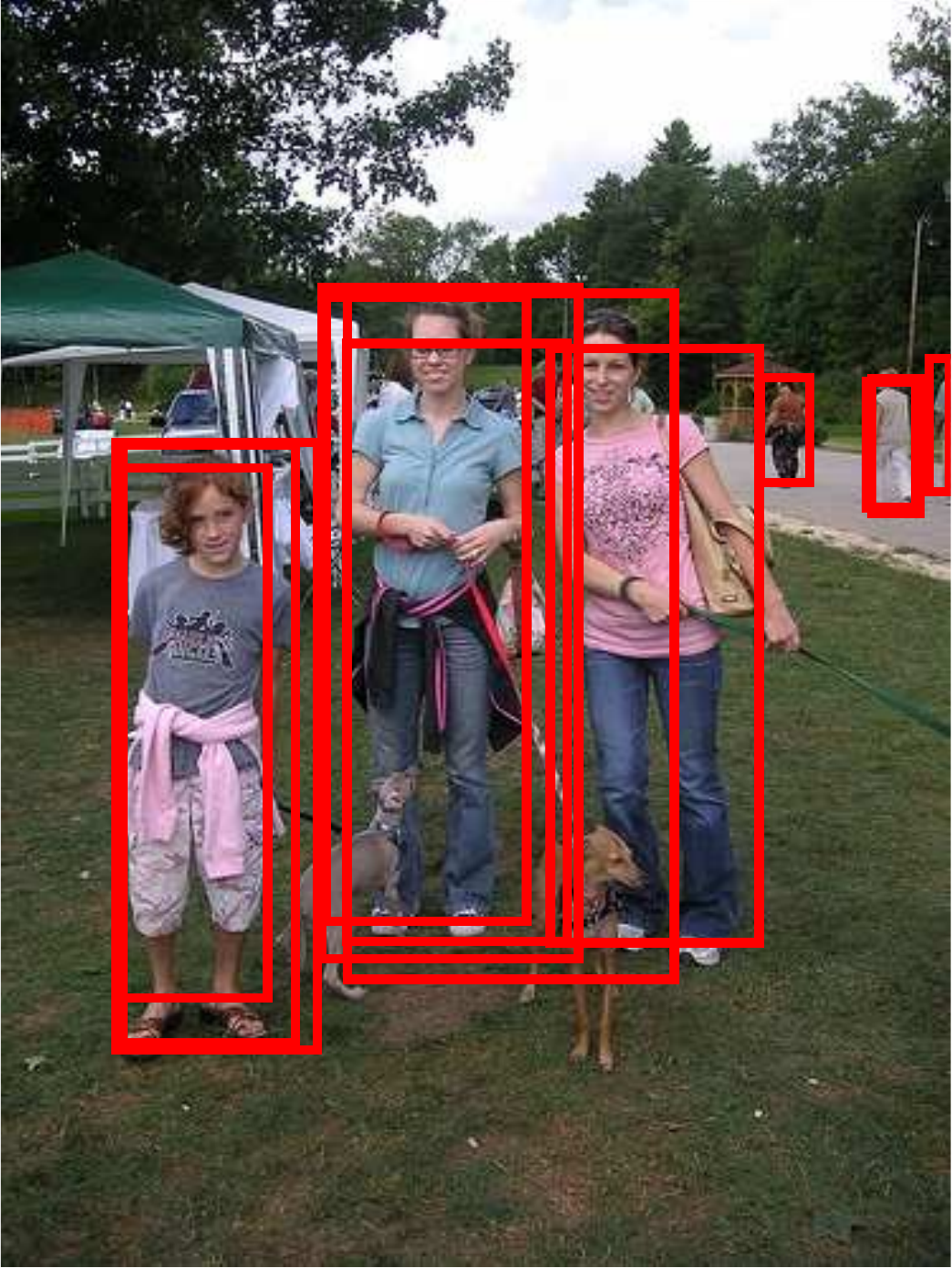}&
\includegraphics[width=0.17\linewidth]{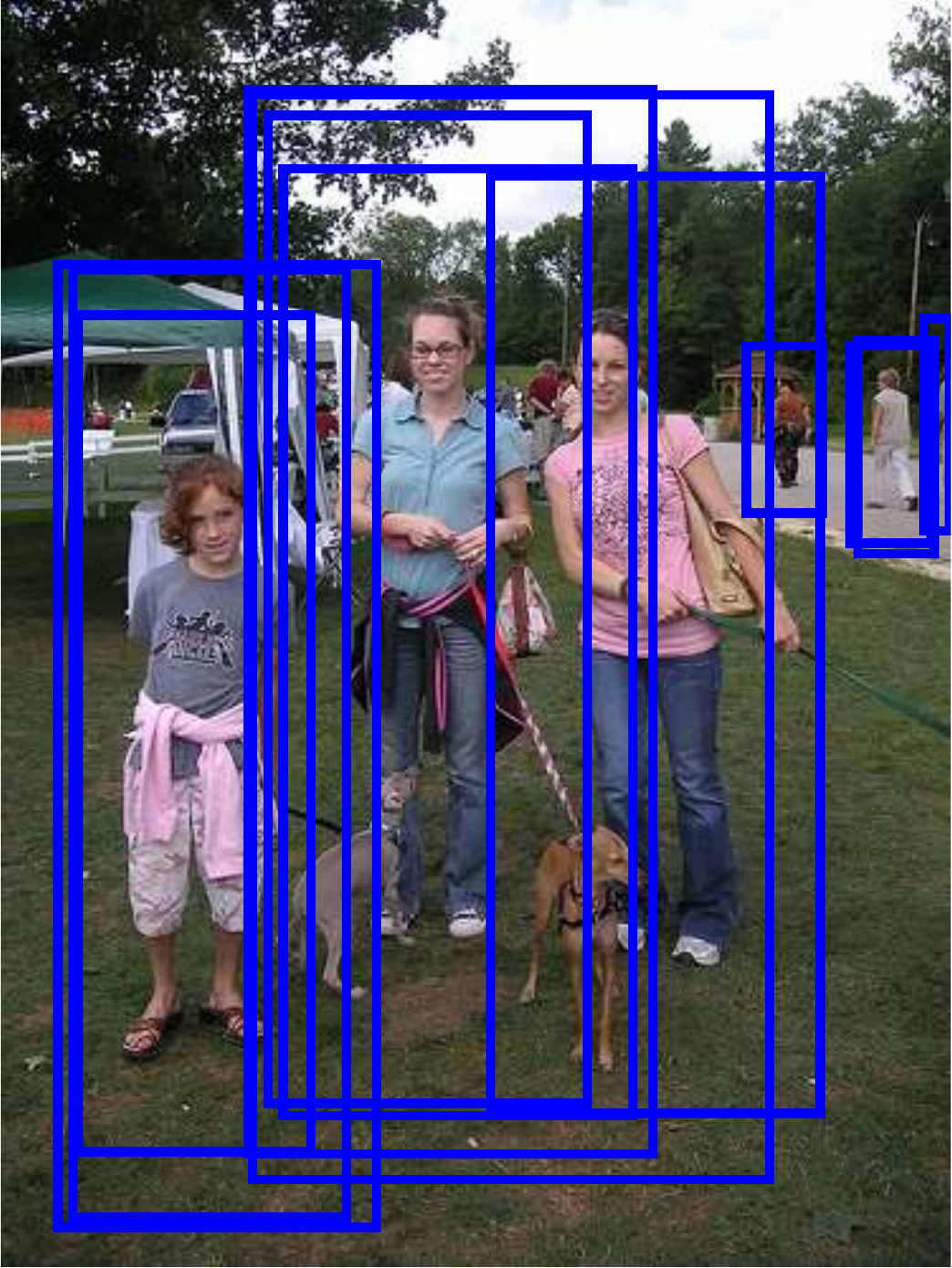}&
\includegraphics[width=0.17\linewidth]{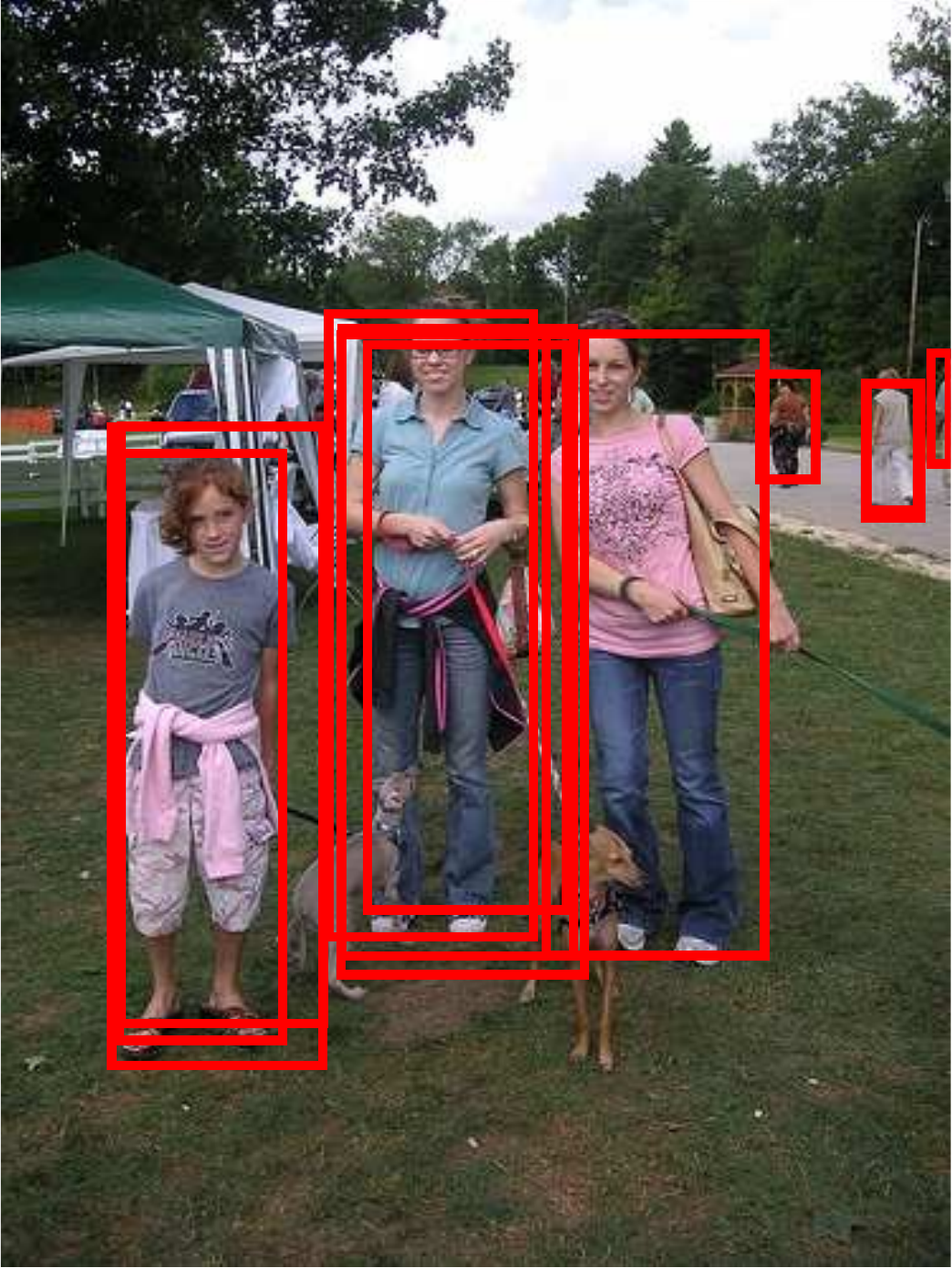}&
\includegraphics[width=0.17\linewidth]{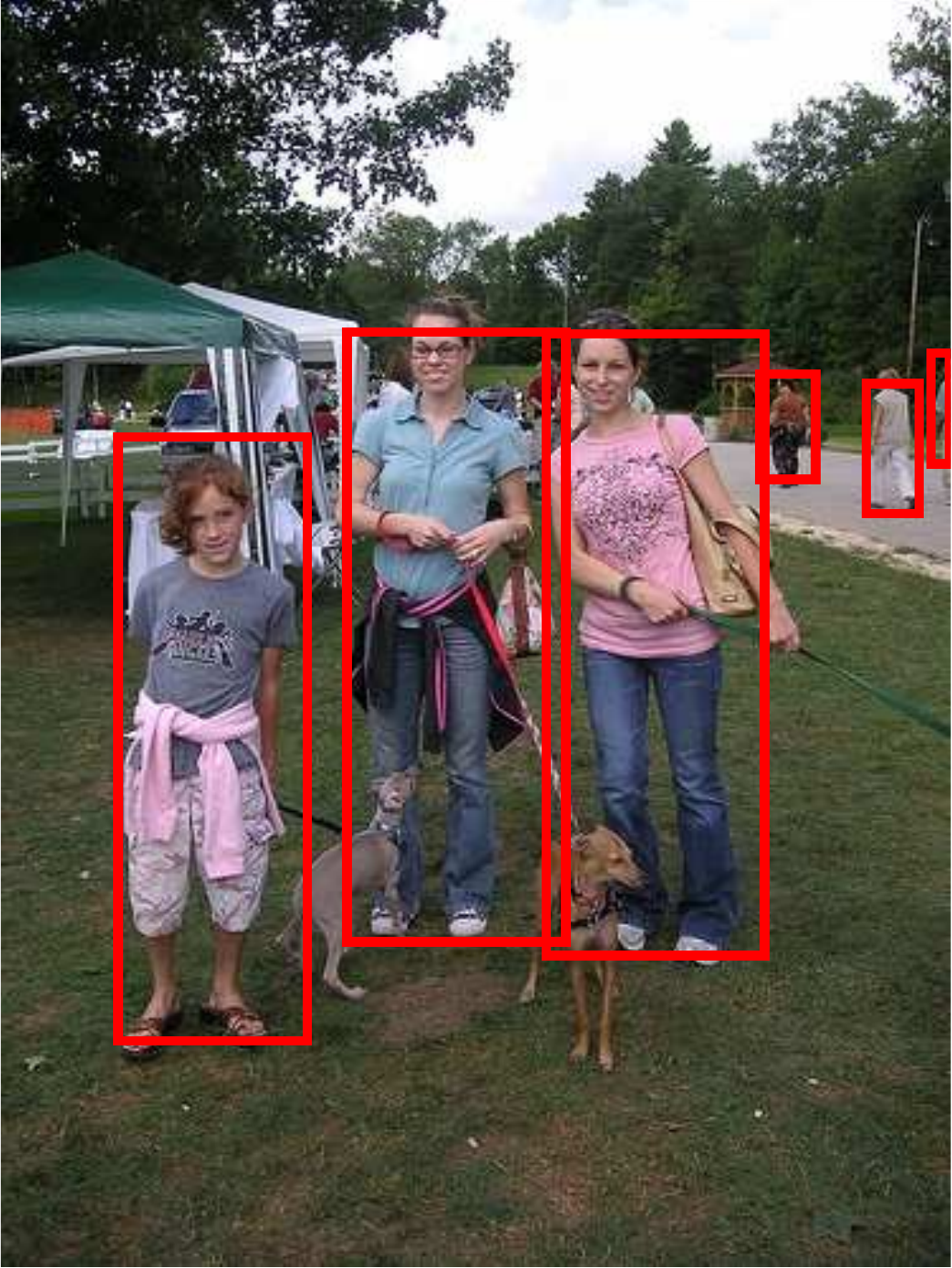}\\
(a) Initial detections.&(b) Initial merge.&(c) Rescale.&(d) Re-detections.&(e) Final merge.
\end{tabular}
\end{center}
\caption{Intermediate results during our detection procedure. (a) shows initial detection results from the initial glances. The boxes are merged as in (b) with an IoU $\mu_0$. To refine the results, the boxes are rescaled with a factor $\beta$ as in (c) and shrink again as in (d). The resulting boxes are finally merged with an IoU $\mu_1$.}
\label{fig:anet:method:refine}
\end{figure*}

Each initial glance is fed to the attention model recurrently until this meets $(\bullet_{\text{TL}},\bullet_{\text{BR}})$ or $(\times_{\text{TL}},\times_{\text{BR}})$. The first image in \Fref{fig:anet:method:refine} shows a real example of the initial detection. The bounding-boxes are merged to a decreased number by a single-linkage clustering; a group of bounding-boxes satisfying a minimum IoU $\mu_{0}$ are averaged into one with their scores of \Eref{eq:anet:method:score}.

To refine the result, \cite{felzenszwalb2010object,girshick2014rich} conduct a post bounding-box regression, which re-localizes the bounding-boxes. This is a linear regression model which maps a feature of the bounding-box to a new one. In our case, we can employ the attention model again for this refinement step. We simply rescale each bounding-box in \Fref{fig:anet:method:refine}-(b) to a new region with a rescaling factor of $\beta$ as shown in \Fref{fig:anet:method:refine}-(c). These reinitialized regions are fed to the attention model again and result in new bounding-boxes as shown \Fref{fig:anet:method:refine}-(d). This re-detection procedure gives us one more chance to reject false positives as well as fine localization. These bounding-boxes are merged again to final results with an IoU $\mu_{1}$.

\section{Experiments}
\label{sec:anet:exp}
In this section, we perform human detection task on public datasets to comprehensively verify the strength of our detection mechanism. Among a wide range of object classes, it is beyond question that the class ``human'' has taken center stage in object detection for decades due to its broad applications. Nonetheless, human detection on uncontrolled natural images is still challenging due to the self-occlusions, diverse poses and clothes.

\subsection{Datasets}
\label{sec:anet:exp:dataset}
For human detection task, we choose PASCAL VOC 2007 and 2012 \cite{Everingham10} since these are composed of user taken web photos from Flickr so the image condition is completely uncontrolled. A lot of human instances in these sets are severely occluded, truncated and overlapped with diverse pose variations and scales. Each of PASCAL VOC 2007 and 2012 include 10K and 23K images of 20 object classes and equally divide into a \texttt{trainval} set and a \texttt{test} set. Following the standard protocol used in the previous human detection researches over these sets, we use \texttt{trainval} images for training, and report an average precision (AP) on \texttt{test} set. For PASCAL VOC 2012, we submit our results to the evaluation server and receive the AP.

\subsection{Base Network}
\label{sec:anet:exp:basenet}
We choose VGG-M~\cite{chatfield2014return} and VGG-16~\cite{simonyan2014very} as the base networks for the attention model. VGG-M, designed by Chatfield~\etal~\cite{chatfield2014return}, is a variant of AlexNet~\cite{krizhevsky2012imagenet} with small modifications. The stride and filter of the first layer are smaller than those of AlexNet but the stride at the second convolution layer is larger. This model is also composed of 8 convolution layers. We adopt this model due to its lower Top-5 error (16.1\%) than AlexNet on the ILSVRC classification without significant increase in computation. VGG-16, designed by Simonyan and Zisserman~\cite{simonyan2014very}, is a much deeper network composed of 16 convolution layers with smaller filter sizes. This single model shows 9.9\% Top-5 error on the ILSVRC classification. 

The PASCAL VOC series is too small to learn this large model. Thus we initialize the base network with pre-trained weights on ILSVRC CLS-LOC dataset \cite{russakovsky2015imagenet} and fine-tune the model for our target task. For each model, we pick out all the pre-trained layers except for the last classification layer and stack our action layers on top of those.

\subsection{Parameters} 
\label{sec:anet:exp:param}
We follow the fine-tuning technique in \cite{agrawal2014analyzing} to train the attention model; an initial learning rate of the pre-trained weights is 0.001 whereas that of the randomly initialized weights is 0.01. When the learning curve is saturated, we decrease the learning rates 0.1-times. For the inference stage, we set the size of each movement action $l$ to 30 pixels for initial detection and 15 pixels for refinement. The number of scaling factors $K$ is 24 with a margin factor $\alpha$ of $\sqrt{6}$. The merging parameters ($\mu_{0}$, $\mu_{1}$) are (1.0, 0.5). The rescaling factor $\beta$ is 3.0.

\subsection{Results and Analysis}
\begin{table}
\begin{center}
\setlength{\tabcolsep}{1.7pt}
\begin{tabular}{|l|c|c|c|}
\hline
Methods&Class&\bigcell{c}{VOC'07\\Person\\AP(\%)}&\bigcell{c}{VOC'12\\Person\\AP(\%)}\\
\hline\hline
AttentionNet&Single&\textbf{71.4}&\textbf{72.8}\\
AttentionNet-Refine&Single&69.9&72.4\\
\hline\hline
Person R-CNN + BB-Regression&Single&59.7&-\\
Person R-CNN + BB-Regression$\times 2$&Single&\textbf{59.8}&-\\
Person R-CNN + BB-Regression$\times 3$&Single&59.7&-\\
\hline\hline
DPM~\cite{felzenszwalb2010object}&Multi&41.9&-\\
Poselets~\cite{bourdev2010detecting}&Single&46.9&-\\
$k$-Poselets\cite{gkioxari2014using}&Single&45.6&-\\
G-P, HOG-III~\cite{jiang2015combination}&Single&55.5&57.0\\
Poselets (AlexNet)~\cite{bourdev2014deep}&Single&59.3&58.7\\
Regression (AlexNet)~\cite{szegedy2013deep}&Multi&26.2&-\\
DeepMultiBox (AlexNet)~\cite{erhan2014scalable}&Multi&37.5&-\\
R-CNN (AlexNet)~\cite{girshick2014rich}&Multi&58.7&57.8\\
R-CNN + SPP (ZFNet)~\cite{he2015spatial}&Multi&57.6&-\\
Q-Learning (ZFNet)~\cite{caicedo2015active}&Single&45.7&-\\
Q-Learning (AlexNet)~\cite{Mathe_2016_CVPR}&Multi&-&18.7\\
Our Multi-class AttentionNet (VGG-M)&Multi&\textbf{72.0}&\textbf{72.4}\\
\hline
\end{tabular}
\;\\(a) Results with 8-layered networks.\\\;
\begin{tabular}{|l|c|c|c|}
\hline
Methods&Class&\bigcell{c}{VOC'07\\Person\\AP(\%)}&\bigcell{c}{VOC'12\\Person\\AP(\%)}\\
\hline\hline
AttentionNet&Single&\textbf{77.6}&\textbf{75.4}\\
AttentionNet-Refine&Single&77.4&75.3\\
\hline\hline
AttentionNet + Faster R-CNN&Single&81.5&79.9\\
AttentionNet-Refine + Faster R-CNN&Single&\textbf{82.1}&\textbf{81.4}\\
\hline\hline
R-CNN (VGG-16)~\cite{girshick2014rich}&Multi&64.2&65.2\\
Fast R-CNN (VGG-16)~\cite{girshick2015fast}&Multi&69.0&69.8\\
Faster R-CNN (VGG-16)~\cite{ren2015faster}&Multi&76.3&77.4\\
MR-CNN + S-CNN + Loc. (VGG-16)~\cite{gidaris2015object}&Multi&74.9&76.4\\
YOLO (26 layers)~\cite{Redmon_2016_CVPR}&Multi&-&63.5\textbf{*}\\
SSD (VGG-16)~\cite{liu2015ssd}&Multi&76.6&\textbf{83.3}\textbf{*}\\
Our Multi-class AttentionNet (VGG-16)&Multi&\textbf{80.1}&79.2\\
\hline
\end{tabular}
\textbf{*}trained on a superset of \texttt{trainval'07+test'07+trainval'12}.\\
(b) Results with networks composed of more than 16 layers.\\
\end{center}
\caption{Human detection performances on PASCAL VOC 2007/2012 \texttt{test} set.}
\label{tab:anet:exp}
\end{table}

We evaluate our method by the human detection task on PASCAL VOC 2007 and 2012. The results and comparisons with 8-layered networks (e.g. VGG-M~\cite{chatfield2014return}, AlexNet~\cite{krizhevsky2012imagenet}, ZFNet~\cite{zeiler2014visualizing}) are shown in \Tref{tab:anet:exp}-(a), and those with networks composed of more than 16 layers (e.g. VGG-16~\cite{simonyan2014very}, YOLO~\cite{Redmon_2016_CVPR}) are shown in \Tref{tab:anet:exp}-(b). When our attention model is based on VGG-M, our method achieves 71.4\% and 72.8\% for each dataset without the refinement step of \Fref{fig:anet:method:refine}-(c$\sim$e). After the refinement step, marked as ``AttentionNet-Refine'', we achieve slightly worse performances of 69.9\% and 72.4\%. When our attention model is equipped with the VGG-16 model, the performances significantly increase to 77.6\% and 75.4\% for each dataset. After the refinement step, the performances are also slightly decreased to 77.4\% and 75.3\%. This single class detection has benefit from the refinement step but we observe that this improves the multi-class detection to be presented in \Sref{sec:manet:method}.

We can reinterpret our method as a regression through iterative classifications, so we compare ours with the detection-by-regression methods such as bounding box regression. To do so, we train a ``Person R-CNN'' and a bounding box regressor ``BB-Regression'' by using the official code\footnote{https://github.com/rbgirshick/rcnn} provided by the R-CNN authors \cite{girshick2014rich}. Only images of ``person'' class are used as positives while the other images are used as backgrounds, to make the comparison completely fair. The initial detection boxes from R-CNN are given to the bounding box regressor then the boxes are re-localized. This method shows 59.7\% as shown in the second block of \Tref{tab:anet:exp}-(a). As our method which repeats actions, we also repeat the bounding-box regression in R-CNN, which is noted by ``BB-Regression$\times1,2$''. However, the improvement is negligible: +0.1\% and +0.0\% for the second and third iterations. Our method beats these approaches with a large margin more than +10\% and these results verify the effectiveness of the iterative classifications as a regression method. 

There are more detection-by-regression methods \cite{szegedy2013deep,erhan2014scalable} in which the network is trained to produce a target object mask \cite{szegedy2013deep} or bounding-box coordinates \cite{erhan2014scalable} for the purpose of class-agnostic object proposals. Still, our method clearly outperforms these methods as shown in \Tref{tab:anet:exp}-(a). Quite recently, \cite{Redmon_2016_CVPR,liu2015ssd} estimate a bounding-box for each grid cell of a convolution feature map, so all the outputs are obtained in a single feed-forward while performing in real-time. The performances of these methods are summarized in \Tref{tab:anet:exp}-(b). YOLO~\cite{Redmon_2016_CVPR} with 26 convolution layers shows 63.5\% which is much lower than ours. SSD~\cite{liu2015ssd} based on VGG-16 achieves the state-of-the-art performance of 83.3\%, however, this model is trained with a \textit{superset} of \texttt{trainval'07+test'07+trainval'12}. These results also verify the benefit of our iterative classifications compared to these regression approaches.

Poselets-based methods \cite{bourdev2010detecting,gkioxari2014using,bourdev2014deep} are related to ours since these methods are limited to a single object class (e.g. human). Among them, \cite{bourdev2014deep} is a deep learning approach which uses an 8-layered network (AlexNet) like ours. However, our method significantly outperforms all these approaches. Through these results, we can expect that our method can be successfully extended to multiple classes, since we do not use a human-specific model to detect humans.

R-CNN~\cite{girshick2014rich} and its advanced variants~\cite{girshick2015fast,ren2015faster} have been the most successful detection method so far. These bottom-up methods contrast with our top-down approach. Let us compare our method with R-CNN using 8 layers in \Tref{tab:anet:exp}-(a). In VOC 2007, both \cite{girshick2014rich} and \cite{he2015spatial} show around 58\% performance, and our method outperforms them with 71.4\% performance. For a fair comparison, we train another R-CNN that only detect the ``person'' class noted by ``Person R-CNN'' but the result is similar (59.7\%). Our result from the 8-layered network is even significantly better than that of R-CNN with a 16-layered network in \Tref{tab:anet:exp}-(b). The performances of VGG-16 based Fast R-CNN~\cite{girshick2015fast} and Faster R-CNN~\cite{ren2015faster} are summarized in \Tref{tab:anet:exp}-(b). Fast R-CNN shows 69.0\% and 69.8\% performances while our method beats them with 77.6\% and 75.4\% performances in VOC 2007 and 2012 respectively. Our method is competitive to Faster R-CNN which shows 76.3\% and 77.4\% performances. 

These two methods, Fast R-CNN and Faster R-CNN, are faster than ours since they are feed-forward while ours is recurrent. Also, they are more advantageous in terms of recall because they only take visible regions into account. In contrast, our method has strong properties driven from scene-level contexts. Thus, we can expect a large performance improvement when we mix these complementary methods. We combine the boxes from Faster R-CNN with our boxes. We rescale our scores, and merge all the boxes and scores with an IoU of 0.5. The results are reported in the second block of \Tref{tab:anet:exp}-(b). We achieve the significantly boosted performances; 82.1\% from 77.4\% in VOC 2007 and 81.4\% from 75.3\% in VOC 2012. Through this experiment, we expect that our research on the top-down approach can contribute to a future hybrid model taking the advantages from both approaches.

There has been two top-down approaches for object detection~\cite{caicedo2015active,Mathe_2016_CVPR}, which adopt reinforcement Q learning~\cite{sutton1998reinforcement} to train their agents with rewards. The performances of these methods are 45.7\% in VOC 2007 and 18.7\% in VOC 2012 respectively, which are far from the state-of-the-art performance, as shown in \Tref{tab:anet:exp}-(a). In contrast, our actions are designed to be optimally chosen at any state so we can train our attention model with a softmax loss, and achieve much superior performances.

We perform extra experiments regarding other behaviors of our detection mechanism and analyze the results in \Sref{sec:manet:exp} with a multi-class attention model.


\section{Multi-Class Detection Mechanism}
\label{sec:manet:method}
In this section, we generalize the attention model that has been specified for an object class to multiple object classes. We first modify the attention model to a multi-class version in \Sref{sec:manet:method:model}. We then explain the initial glance mining in \Sref{sec:manet:method:glance} followed by the remaining detection procedures in \Sref{sec:manet:method:refine}. We finally present the training method in \Sref{sec:manet:method:training}.

\subsection{Attention model}
\label{sec:manet:method:model}
\begin{figure*}[t]
\begin{center}
\includegraphics[width=0.75\linewidth]{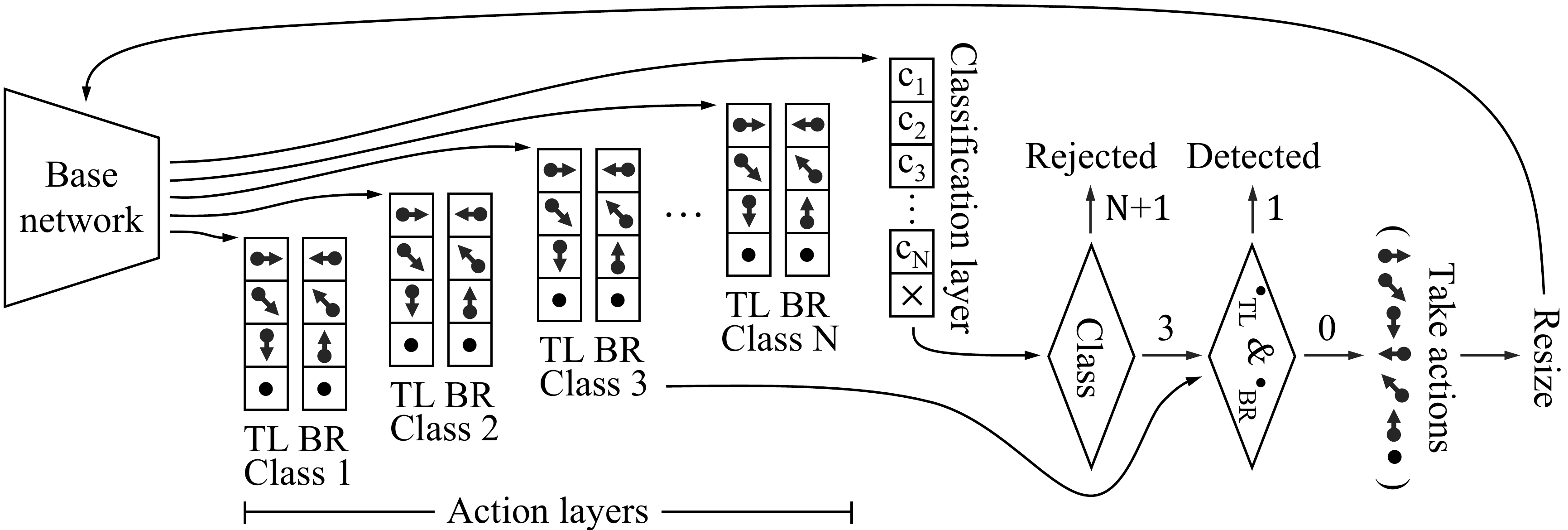}
\end{center}
\caption{An attention model extended to multiple object classes. A pair of TL and BR action layers are defined for each class to provide class-specific actions. A classification layer is also defined to recognize the class of an input object.}
\label{fig:manet:method:model}
\end{figure*}

In the class-specific attention model, a pair of action layers is specified for a single object class. To make the model can provide actions regarding multiple classes, we define the action layers for each class and parallelize them on top of a base network. This model is illustrated in \Fref{fig:manet:method:model}. If we have $N$ classes, the end of the model is composed of $N$ pairs of action layers. Given an input, this model always produces $N$ class-specific action pairs. However, only with these, we can not determine which action pair should be chosen for this input since the object class of the input is unknown. Thus, we define an extra classification layer which recognizes the input object class so that we are able to choose an action pair. The classification layer produces a ($N$+$1$)-dimensional vector composed of a background score and the $N$ class scores. Since the classification layer tells us whether an input is background or not, we remove ``reject $\times$'' actions from all the action layers. Thus, each action layer produces a 4-dimensional vector composed of the 3 movement action scores and a stop action score.

The multi-class detection mechanism with this attention model proceeds as follows. From an input, we obtain classification scores $\mathbf{y}_\text{cls}$=$[y_1\cdots y_{N+1}]$ and action scores $\left\{\mathbf{y}_{\text{TL}c},\mathbf{y}_{\text{BR}c}\;|\;c=1\cdots N\right\}$. Here the action scores for each class are $\mathbf{y}_{\text{TL}c}$=$[y^{\rightarrow}_{\text{TL}c},y^{\searrow}_{\text{TL}c},y^{\downarrow}_{\text{TL}c},y^{\bullet}_{\text{TL}c}]$ and $\mathbf{y}_{\text{BR}c}$=$[y^{\leftarrow}_{\text{BR}c},y^{\nwarrow}_{\text{BR}c},y^{\uparrow}_{\text{BR}c},y^{\bullet}_{\text{BR}c}]$. If the input is predicted as a background class $\hat{c}$=$N$+1, the input is rejected. If not, we choose the predicted actions of the predicted class $\hat{c}$ for both corners and then take the actions. This procedure is repeated until the action predictions for both corners are ($\bullet_{\text{TL}\hat{c}},\bullet_{\text{BR}\hat{c}}$), or the input is rejected with $\hat{c}$=$N$+1.

A detected region is back-projected to a corresponding bounding-box $b$ in the original input image domain. The detection score $s^b$ is also discriminatively defined as
\begin{multline}
s^b=\left(1-\gamma\right)s_{cls}^b+\gamma\left(s^b_{\text{TL}}+s^b_{\text{BR}}\right)\\
\quad \text{s.t.}\quad\quad
\begin{split}
s_{cls}^b&=y_{\hat{c}}-y_{N+1}\\
s^b_{\text{TL}}&=y^{\bullet}_{\text{TL}\hat{c}}-(y^{\rightarrow}_{\text{TL}\hat{c}}+y^{\searrow}_{\text{TL}\hat{c}}+y^{\downarrow}_{\text{TL}\hat{c}})\\
s^b_{\text{BR}}&=y^{\bullet}_{\text{BR}\hat{c}}-(y^{\leftarrow}_{\text{BR}\hat{c}}+y^{\nwarrow}_{\text{BR}\hat{c}}+y^{\uparrow}_{\text{BR}\hat{c}})
\label{eq:score}
\end{split}
\end{multline}
where $\gamma$ is a fusion parameter of the classification and action scores. Here each score $y$ is a value before the softmax normalization.

\subsection{Initial Glance}
\label{sec:manet:method:glance}
To detect multiple instances in an image, we collect regions called initial glances as presented in \Sref{sec:anet:method:glance}. A required condition for a region to be an initial glance is that the region should contain an entire object body with sufficient surrounding contexts. Since we recurrently shrink an initial glance to a target object with movement actions, it is important for the initial glance not to be truncating a target object. To this end, given a lot of candidate regions, we choose regions which satisfy the following conditions as initial glances
\begin{equation}
\hat{c}\neq N+1\;\;\;\text{and}\;\;\;\hat{a}_{\text{TL}\hat{c}}=\searrow_{\text{TL}}\;\;\;\text{and}\;\;\;\hat{a}_{\text{BR}\hat{c}}=\nwarrow_{\text{BR}}
\label{eq:condition}
\end{equation}
which indicate that an initial glance should not be a background, and should not be truncating the target object as shown in \Fref{fig:anet:method:glance}. Here, $\hat{c}$ is a class prediction from the classification layer, and ($\hat{a}_{\text{TL}\hat{c}}$, $\hat{a}_{\text{BR}\hat{c}}$) are the action predictions from the action layers of the predicted class $\hat{c}$. 

To boost speed and recall of the initial glance mining, we feed multi-scale multi-aspect ratio images to our fully convolutional attention model and pick out the initial glances over the resulting action maps as illustrated in \Fref{fig:anet:method:glance-mining}. We also follows the data-driven method to determine the scaling factors which has presented in \Sref{sec:anet:method:glance}. The scaling factors $\{\mathbf{s}_k\;|\;k=1\cdots K\}$ are estimated from all the ground truth bounding boxes regardless of their classes in a training image set.

\subsection{Initial Detection and Refinement}
\label{sec:manet:method:refine}
Start from the initial glances, we iteratively shrink the boundaries with the movement actions $(\hat{a}_{\text{TL}\hat{c}}, \hat{a}_{\text{BR}\hat{c}})$ of the predicted class $\hat{c}$. This procedure is repeated until a region is rejected with $\hat{c}$=$N$+1, or meets the following conditions
\begin{equation}
\hat{c}\neq N+1\;\;\;\text{and}\;\;\;\hat{a}_{\text{TL}\hat{c}}=\bullet_{\text{TL}}\;\;\;\text{and}\;\;\;\hat{a}_{\text{BR}\hat{c}}=\bullet_{\text{BR}}
\label{eq:stop}
\end{equation}
which indicate that the final region is a foreground class and terminates with the stop actions at both corners. We back-project the final regions to the corresponding boxes in the original input image domain, and these boxes are then merged with an initial IoU $\mu_0$. As a refinement step, we rescale the initial detection boxes with a rescaling factor $\beta$, and shrink them again until they terminate with the stop actions. We finally merge the detection boxes with a final IoU $\mu_1$. \Fref{fig:anet:method:refine} shows real examples for these procedures.

\subsection{Training}
\label{sec:manet:method:training}
To make the multi-class attention model recognize the class of an object and proper actions for that object, we train the model with random regions which evenly cover the various object classes, their required actions and backgrounds. Because ground-truths including bounding-boxes and classes are given, we can automatically determine the optimal action label for each corner of a random region. We follows the method in \Fref{fig:anet:method:training} to generate random regions. When we compose a mini-batch for training, we select the object regions and the background regions with an equal probability as described in \Sref{sec:anet:method:training}. 

Given a region with a class label $t_{cls}$ and a pair of action labels ($t_{\text{TL}},t_{\text{BR}}$), we can define three log-softmax losses at a classification layer and the two action layers of the class. Zero-losses are given to the action layers of the other classes. We define a final loss $\ell$ by combining a classification loss $\ell_{\text{cls}}$ and the two action losses ($\ell_{\text{TL}}, \ell_{\text{BR}}$) such as
\begin{multline}
\ell=\lambda\cdot \ell_{\text{cls}}+\frac{1-\lambda}{2}\cdot\left(\ell_{\text{TL}}+\ell_{\text{BR}}\right)
\quad \text{s.t.}\\\\
\begin{split}
&\ell_{\text{cls}}=\ell_{\text{softmax}}\left(\mathbf{y}_{\text{cls}}, t_{\text{cls}}\right)\\
&\ell_{\text{TL}}=\sum_{c=1}^{N}\mathbb{1}(c,t_{\text{cls}})\cdot\ell_{\text{softmax}}\left(\mathbf{y}_{\text{TL}c}, t_{\text{TL}}\right)\\
&\ell_{\text{BR}}=\sum_{c=1}^{N}\mathbb{1}(c,t_{\text{cls}})\cdot\ell_{\text{softmax}}\left(\mathbf{y}_{\text{BR}c}, t_{\text{BR}}\right)\\
&\ell_{\text{softmax}}\left(\mathbf{y}, t\right)=-y_t+\log\sum_{i}e^{y_i}
\end{split}
\label{eq:multi-loss}
\end{multline}
where $\lambda$ is a constant from a range $[0, 1]$, and $\mathbb{1}(c,t)$ is 1 if $c$ equals to $t$ but 0 for otherwise. If a region is a background $t_{cls}$=$N$+1, the losses from all the action layers become zero.

\section{Multi-Class Experiments}
\label{sec:manet:exp}
In this section, we perform multi-class object detection task on public datasets to comprehensively verify our detection mechanism extended to multiple classes.

\subsection{Datasets}
For primary evaluation, we choose PASCAL VOC 2007 and 2012 \cite{everingham2010pascal} dataset. We follows the standard protocol of these sets as described in \Sref{sec:anet:exp:dataset}. Also, we select ILSVRC CLS-LOC dataset \cite{russakovsky2015imagenet} to verify our detection capability for large number of classes with large-scale data. This dataset includes images of 1,000 object classes and divide into 1.3M training images, 50K validation images and 100K test images. All the images are annotated with object classes but only a part of the training set, 524K images, contain bounding box annotations. We use these 524K images to train the attention model. To evaluate localization, we submitted the localization results on the \texttt{test} set to ILSVRC'15 evaluation server and received the Top-5 LOC error. Top-5 LOC error reflects classification and localization errors at once from top-5 predictions for each image. This metric is defined in Sec. 4.2 of \cite{russakovsky2015imagenet}.

\subsection{Base Network}
We also use VGG-M~\cite{chatfield2014return} and VGG-16~\cite{simonyan2014very} for the PASCAL VOC series in the same way presented in \Sref{sec:anet:exp:basenet}. For ILSVRC CLS-LOC dataset, we use GoogLeNet designed by Szegedy~\etal~\cite{szegedy2015going}. GoogLeNet is also a deep model that has 22 convolution layers but much faster than VGG-16 while showing a comparative performance of 12.9\% Top-5 error with a single model. We choose this model to speed up training over the large-scale data. Before training, we pick out all the pre-trained layers of GoogLeNet except for the last classification layer and stack our action layers on top of those.

\subsection{Parameters} 
All the parameters necessary for training and inference are equal to those of the single class detection mechanism, which has been presented in \Sref{sec:anet:exp:param}, except for the following; we use the number of scaling factors $K$ of 7 and the rescaling factor $\beta$ of 8.0 for the evaluation with ILSVRC CLS-LOC dataset. The loss fusion parameter $\lambda$ and the score fusion parameter $\gamma$ are equally set to 0.5.

\subsection{Results and analysis} 
\begin{table*}
\setlength{\tabcolsep}{1.4pt}
\footnotesize
\begin{center}
\begin{tabular}{|l|c|c|c|c|c|c|c|c|c|c|c|c|c|c|c|c|c|c|c|c|c|c|}\hline
Method&\bigcell{c}{Base net}&plane&bike&bird&boat&bottle&bus&car&cat&chair&cow&table&dog&horse&motor&person&plant&sheep&sofa&train&tv&\bigcell{c}{mAP\\(\%)}\\\hline\hline
AttentionNet&VGG-M&74.6&71.0&56.8&52.0&47.8&72.1&79.9&67.8&35.9&66.3&49.4&64.5&72.9&70.0&71.4&42.4&64.6&53.5&70.1&62.6&62.3\\
AttentionNet-Refine&VGG-M&76.5&73.0&57.8&52.3&51.8&71.9&80.0&67.3&37.1&67.8&49.4&65.9&72.3&69.0&72.0&43.2&66.3&53.6&71.9&66.1&\textbf{63.3}\\\hline\hline
AttentionNet&VGG-16&77.4&77.1&70.3&57.4&58.9&76.7&85.5&75.4&45.0&80.1&61.3&76.2&78.3&75.3&78.2&47.7&73.4&63.3&72.5&71.6&70.1\\
AttentionNet-Refine&VGG-16&79.1&77.6&70.2&58.0&60.0&75.8&85.5&75.9&47.6&79.7&61.6&76.9&78.6&76.0&80.1&47.0&73.9&64.3&74.1&72.5&\textbf{70.7}\\\hline
\end{tabular}
\;\\(a) Our results on PASCAL VOC 2007 \texttt{test} set. All the method use \texttt{trainval'07} set for training.\\\;
\begin{tabular}{|l|c|c|c|c|c|c|c|c|c|c|c|c|c|c|c|c|c|c|c|c|c|c|}\hline
Method&\bigcell{c}{Base net}&plane&bike&bird&boat&bottle&bus&car&cat&chair&cow&table&dog&horse&motor&person&plant&sheep&sofa&train&tv&\bigcell{c}{mAP\\(\%)}\\\hline\hline
AttentionNet&VGG-M&77.7&66.1&61.5&44.8&46.8&70.9&72.0&73.7&32.6&59.4&40.0&71.4&64.8&70.8&70.6&38.5&69.5&38.8&70.0&56.8&59.8\\
AttentionNet-Refine&VGG-M&77.8&66.4&60.9&44.9&48.5&71.8&72.3&75.3&34.5&59.4&40.4&71.5&65.5&71.3&72.4&38.3&70.9&40.9&70.1&59.3&\textbf{60.6}\\\hline\hline
AttentionNet&VGG-16&77.0&69.4&64.1&51.7&53.9&73.1&76.3&77.0&39.3&69.4&42.2&74.9&72.6&73.5&76.2&46.5&73.2&44.1&73.2&61.1&64.4\\
AttentionNet-Refine&VGG-16&79.1&68.9&65.5&52.3&55.9&73.5&76.5&79.1&42.3&70.9&42.7&76.8&73.9&73.3&79.2&48.1&74.1&44.9&73.7&62.2&\textbf{65.6}\\\hline
\end{tabular}
\;\\(b) Our results on PASCAL VOC 2012 \texttt{test} set. All the method use \texttt{trainval'12} set for training.\\
\end{center}
\caption{Our multi-class detection performances on (a) PASCAL VOC 2007 \texttt{test} set and (b) PASCAL VOC 2012 \texttt{test} set.}
\label{tab:manet:exp:ours}
\end{table*}

We evaluate our multi-class detection model with PASCAL VOC 2007 and 2012. The per-class APs(\%) of our method at these datasets are listed in \Tref{tab:manet:exp:ours}. When the base network of our multi-class attention model is VGG-M, our method achieves 62.3\% and 59.8\% for each dataset without the refinement step of \Fref{fig:anet:method:refine}-(c$\sim$e). After the refinement step, noted by ``AttentionNet-Refine'', the performances are improved to 63.3\% and 60.6\%. Different from the single class attention model in \Tref{tab:anet:exp}, the refinement step gives a non-negligible benefit coming from re-localizations. Note, our multi-class attention model is reused for the refinement. When the model size increases to the VGG-16 model, performances also significantly increase to 70.1\% and 64.4\% for each dataset. With the refinement step, the performances slightly increase to 70.7\% and 65.6\%.

\begin{table}
\begin{center}
\setlength{\tabcolsep}{1.7pt}
\begin{tabular}{|l|l|c|c|c|c|c|}
\hline
Base net&Motion size $l$ (Pixels)&5&10&15&30&50\\
\hline\hline
\multirow{2}{*}{VGG-M}&Mean number of movements&23.5&12.9&9.3&5.5&4.0\\
&mAP (\%) of initial detection&60.6&62.6&\textbf{62.9}&62.3&57.6\\\hline\hline
\multirow{2}{*}{VGG-16}&Mean number of movements&26.5&14.3&10.2&6.0&4.3\\
&mAP (\%) of initial detection&69.3&70.4&\textbf{71.0}&70.1&62.5\\
\hline
\end{tabular}
\end{center}
\caption{Initial detection performances and mean number of movements per instance as the motion size increases in PASCAL VOC 2007 \texttt{test} set.}
\label{tab:manet:exp:dvs}
\end{table}

As we have presented in \Sref{sec:anet:model}, the attention model tells each corner which direction to move. The corner then moves $l$ pixels along that direction. The input region is always resized to 224$\times$224 size and the size of a movement $l$ is constant, so the movement in the original image domain becomes smaller then that of the previous stage, as shown in \Fref{fig:teaser}. Let us take a look at how performance varies with the size of the movement $l$ and how many movements are required to detect an object instance. \Tref{tab:manet:exp:dvs} is summarizing the results. The smaller size of movement requires more movements to detect an instance. We expected the accuracy to decrease as the size of movement increases. However, in reality, the accuracy rather increases, and then it starts to decrease later. The reason is because the more iteration, the more chance the attention model has of making false negative decisions $\{\times_\text{TL},\times_\text{BR}\}$. The results in both base networks show a similar tendency. When the size of movement is 15 pixels, our method shows the best accuracy of 62.9\% and 71.0\% in each base network. These results beat Faster R-CNN~\cite{ren2015faster}, which show 59.9\% with ZFNet and 69.9\% with VGG-16, by margins of 3.0\% and 1.1\%, respectively. However, because the size of 30 pixels requires much less number of movements than that of 15 pixels but shows good performances, we set the size to 30 pixels for the initial detection in all the other experiments.

\begin{table}
\begin{center}
\setlength{\tabcolsep}{1.7pt}
\begin{tabular}{|l|c|c|}
\hline
Methods&\bigcell{c}{VOC'07\\mAP(\%)}&\bigcell{c}{VOC'12\\mAP(\%)}\\
\hline\hline
AttentionNet&62.3&59.8\\
AttentionNet-Refine&\textbf{63.3}&\textbf{60.6}\\
\hline\hline
Regression (AlexNet)~\cite{szegedy2013deep}&30.5&-\\
DeepMultiBox (AlexNet)~\cite{erhan2014scalable}&29.2&-\\
R-CNN (AlexNet)~\cite{girshick2014rich}&58.5&\textbf{53.3}\\
R-CNN + SPP (ZFNet)~\cite{he2015spatial}&59.2&-\\
Fast R-CNN (CaffeNet)~\cite{girshick2015fast}&58.4&-\\
Faster R-CNN (ZFNet)~\cite{ren2015faster}&\textbf{59.9}&-\\
Q-Learning (ZFNet)~\cite{caicedo2015active}&46.1&-\\
Q-Learning (AlexNet)~\cite{Mathe_2016_CVPR}&-&27.0\\
\hline
\end{tabular}
\;\\(a) Comparisons with 8-layered networks.\\\;
\begin{tabular}{|l|c|c|c|}
\hline
Methods&\bigcell{c}{VOC'07\\mAP(\%)}&\bigcell{c}{VOC'12\\mAP(\%)}\\
\hline\hline
AttentionNet&70.1&64.4\\
AttentionNet-Refine&\textbf{70.7}&\textbf{65.6}\\
\hline\hline
AttentionNet + Faster R-CNN&74.2&71.0\\
AttentionNet-Refine + Faster R-CNN&\textbf{75.8}&\textbf{71.5}\\
\hline\hline
R-CNN (VGG-16)~\cite{girshick2014rich}&66.0&62.4\\
Fast R-CNN (VGG-16)~\cite{girshick2015fast}&66.9&65.7\\
Faster R-CNN (VGG-16)~\cite{ren2015faster}&69.9&67.0\\
YOLO (26 layers)~\cite{Redmon_2016_CVPR}&-&57.9\textbf{*}\\
SSD (VGG-16)~\cite{liu2015ssd}&71.6&\textbf{74.9}\textbf{*}\\
MR-CNN + S-CNN + Loc. (VGG-16)~\cite{gidaris2015object}&\textbf{74.9}&70.7\\
\hline
\end{tabular}
\\\textbf{*}trained on a superset of \texttt{trainval'07+test'07+trainval'12}.\\
(b) Comparisons with networks composed of more than 16 layers.\\
\end{center}
\caption{Object detection performances on PASCAL VOC 2007/2012 \texttt{test} set.}
\label{tab:manet:exp:comp}
\end{table}

We can regard our method as a box regression by stacked classifications. We therefore compare ours with the traditional detection-by-regression methods such as \cite{szegedy2013deep,erhan2014scalable} of which \Tref{tab:manet:exp:comp}-(a) lists performances. The network of \cite{szegedy2013deep} estimates a object box mask for each sliding window and shows 30.0\% performance. The network of \cite{erhan2014scalable} directly produces box coordinates of class-agnostic object proposals and shows 29.2\% performance. Our method beats these approaches with a large margin more than +30\%. YOLO~\cite{Redmon_2016_CVPR} is a recent detection-by-regression method in which a 26-layered network estimates a bounding-box for each grid cell of a convolutional feature map. As listed in \Tref{tab:manet:exp:comp}-(b), this method shows a much lower performance of 57.9\% than our 65.6\% in VOC 2012 even if this is trained with a \textit{superset} composed of \texttt{trainval'07+test'07+trainval'12}. SSD~\cite{liu2015ssd} is a variant of YOLO but it operates on a multiple scale feature maps, and shows the state-of-the-art performances of 71.6\% and 74.9\% in each dataset. Our performance of 70.1\% is slightly worse than that of SSD in VOC 2007, and SSD outperforms ours with a large margin in VOC 2012 but SSD is also trained with the same \textit{superset} used in YOLO. 

R-CNN~\cite{girshick2014rich} and its advanced variants such as Fast R-CNN\cite{girshick2015fast} and Faster R-CNN~\cite{ren2015faster} are typical bottom-up approaches. Let us compare our top-down approach with them. As shown in \Tref{tab:manet:exp:comp}, the performances of the original R-CNN are significantly worse than ours in general. Our method also beats Fast and Faster R-CNNs in VOC 2007. In VOC 2012, the performance of our method is similar to that of Fast RCNN and lower than that of Faster R-CNN. Our top-down method demonstrated comparable performances compared to these state-of-the-art bottom-up method.

Bottom-up and top-down approaches are complementary to each other. A bottom-up method is feed-forward and efficient. Also, detection from proposals is more advantageous in recall. In contrast, our top-down method is recursive so slower than that but our high-level action cascade is more advantageous in the scene-level context so results in less false positives. Thus, we can expect a large performance improvement when we mix the two complementary approaches. We rescale our scores, and merge all the boxes and their scores with an IoU of 0.5. As shown in the second block of \Tref{tab:manet:exp:comp}-(b), the results are 75.8\% in VOC 2007 and 71.5\% in VOC 2012 with gains of +5.1\% and +5.9\%. These results clearly demonstrate the potential of mixing the top-down and bottom-up methods.

Similar to ours, \cite{caicedo2015active,Mathe_2016_CVPR} also use an agent providing actions but they depend on the reinforcement Q-learning \cite{sutton1998reinforcement} to train that because selecting an optimal action given a state is ambiguous. In VOC 2007, \cite{caicedo2015active} shows 46.1\% which is much worse than our 63.3\%. Also, in VOC 2012, \cite{Mathe_2016_CVPR} shows 27.0\% while ours is 60.6\%. Both of these methods are successful top-down methods using reinforcement learning for detection, but the reinforcement method has a high variance in the gradient of the expected reward so their performances are far from the state-of-the-art methods. In contrast, our top-down mechanism first demonstrates competitive performances compared to the recent bottom-up methods.

\begin{table}
\begin{center}
\setlength{\tabcolsep}{1.7pt}
\begin{tabular}{|l|c|c|c|}
\hline
Methods&\bigcell{c}{IoU=0.5\\mAP(\%)}&\bigcell{c}{IoU=0.7\\mAP(\%)}&\bigcell{c}{mAP(\%)\\drop}\\
\hline\hline
AttentionNet&70.1&49.7&$-$20.4\\
AttentionNet ($l$=15)&\textbf{71.0}&52.3&$-$18.7\\
AttentionNet-Refine&70.7&\textbf{53.1}&$-$\textbf{17.6}\\
\hline\hline
R-CNN NoBBReg (VGG-16)~\cite{girshick2014rich}&60.6&30.8&$-$29.8\\
R-CNN (VGG-16)~\cite{girshick2014rich}&66.0&35.2&$-$30.8\\
R-CNN + Bayesian (VGG-16)~\cite{zhang2015improving}&68.4&43.7&$-$24.7\\
Faster R-CNN (VGG-16)~\cite{ren2015faster}&69.9&46.0&$-$\textbf{23.9}\\
MR-CNN + S-CNN + Loc. (VGG-16)~\cite{gidaris2015object}&\textbf{74.9}&\textbf{48.4}&$-$26.5\\
\hline
\end{tabular}
\end{center}
\caption{Object detection performances with different IoU thresholds (standard 0.5 and strict 0.7) for positive detection in PASCAL VOC 2007 \texttt{test} set. The base network of all the methods is VGG-16.}
\label{tab:manet:exp:iou}
\end{table}

Our detection method has another strength driven from the iterative action classifications. Compared to the bottom-up methods, our box localization is more accurate because it combines several weak actions to estimate a final detection box. \Tref{tab:manet:exp:iou} summarizes the performance comparisons with different IoU thresholds in PASCAL VOC 2007. In our method without the refinement step, the performance drop is -20.4 as the IoU threshold increases from the typical 0.5 to a strict 0.7. When we conduct the refinement step, we observe a smaller performance drop of -17.6 because we have one more chance to correct mis-localizations such as the traditional bounding-box regression R-CNN does. The performance drop of all the other methods is much larger than that of ours, so our method significantly beats all those when IoU threshold is 0.7.

\begin{table}
\begin{center}
\setlength{\tabcolsep}{1.7pt}
\begin{tabular}{|l|l|l|c|c|c|}
\hline
Year&\bigcell{l}{Loc. methods}&\bigcell{l}{Base net\\for loc.}&\bigcell{c}{Model\\depth\\(\#conv)}&\bigcell{c}{Loc. net\\ensemble}&\bigcell{c}{Top-5 LOC err\\(Top-5 CLS err)}\\
\hline\hline
2013&Regression~\cite{sermanet2013overfeat}&OverFeat&4&Yes&0.2988 (0.1568)\\
2014&Regression~\cite{szegedy2015going}&GoogLeNet&22&Yes&0.2644 (0.1483)\\
2014&Regression~\cite{simonyan2014very}&VGG-16&16&Yes&0.2532 (0.0741)\\
2015&Faster R-CNN~\cite{he2016deep}&ResNet&\textbf{152}&Yes&\textbf{0.0902} (\textbf{0.0357})\\
2015&AttentionNet&GoogLeNet&22&\textbf{No}&0.1473 (0.0792)\\
\hline
\end{tabular}
\end{center}
\caption{Performance comparison of 1,000-class object localization on ILSVRC~CLS-LOC \texttt{test} set.}
\label{tab:manet:exp:ilsvrc}
\end{table}

We finally present the localization performance, measured by Top-5 LOC error~\cite{russakovsky2015imagenet}, from a large-scale experiment with ILSVRC CLS-LOC dataset. The comparison between our result\footnote{http://image-net.org/challenges/ilsvrc+mscoco2015} and the existing winning methods is summarized in \Tref{tab:manet:exp:ilsvrc}. All entries being compared with ours are top methods in the ILSVRC localization task. Also, all these are using the bottom-up approaches for localization. \cite{sermanet2013overfeat} is the winning method in 2013, and \cite{szegedy2015going,simonyan2014very} are the first and second places in 2014, respectively. These three methods localize object by the bounding-box regression over dense sliding windows. Compared with these methods, our method shows much lower error of 0.1473 with a large margin of more than 0.1. More recently, He~\etal~\cite{he2016deep} won this challenge in 2015 with Faster R-CNN~\cite{ren2015faster} based on the very deep residual network composed of 152 layers. This method shows a small localization error of 0.0902 which is lower than ours. However, this performance gap mainly comes from the classification performance since the localization error is including the classification error as well. The classification error of the residual network is 0.0357, which is already 2-times smaller than our classification error of 0.0792. Also all the other methods in \Tref{tab:manet:exp:ilsvrc} ensemble multiple localization networks but our localization only depends on \textit{a single multi-class attention model}.

\section{Conclusions}
\label{sec:conclusion}
In this paper, we have proposed a novel method for object detection. 
We adopted a well-studied classification technique for object detection and presented a weak attention model to get high-level actions from that to get closer to a target object. Since we actively explorer an exact bounding-box of a target object in a top-down approach, we do not suffer from the quality of initial object proposals and also take the scene-level context into consideration.

Through this study, we have an important observation that our top-down approach is complementary to the previous state-of-the-art method using a bottom-up approach, therefore combining the two approaches boosts the performance of object detection. Thus, we believe the research on top-down approaches and combining them with the bottom-up methods will likely contribute to the next direction for object detection.



%



\ifCLASSOPTIONcompsoc
  \section*{Acknowledgments}
\else
  \section*{Acknowledgment}
\fi
This work was supported by the Technology Innovation Program (No. 10048320), funded by Korea government (MOTIE). This work was also supported by the National Research Foundation of Korea (No. 2010-0028680), funded by Korea government (MSIP).

\ifCLASSOPTIONcaptionsoff
  \newpage
\fi



\bibliographystyle{IEEEtran}
\bibliography{egbib}
%



%

\begin{IEEEbiography}[{\includegraphics[width=1in,height=1.25in,clip,keepaspectratio]{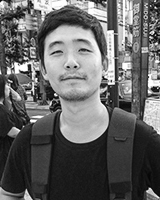}}]{Donggeun Yoo} received BS degree in 2011 and MS degree in 2013 in School of Electrical Engineering, KAIST, South Korea. In present, he is a PhD student in the same school and department. His research interests include representation learning, learning with large-scale data, unsupervised learning and visual recognition. He is a student member of the IEEE.
\end{IEEEbiography}
\begin{IEEEbiography}[{\includegraphics[width=1in,height=1.25in,clip,keepaspectratio]{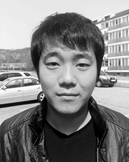}}]{Sunggyun Park} received BS degree in 2010 and MS degree in 2012 in Department of Industrial and Systems Engineering, KAIST, South Korea. In present, he is a PhD student in the same school and department, and also a co-founder and a research scientist at Lunit Inc, South Korea. His research interests include stochastic process for product pricing and machine learning for problems, which are related to operations management and computer vision. He is a student member of the IEEE.
\end{IEEEbiography}
\begin{IEEEbiography}[{\includegraphics[width=1in,height=1.25in,clip,keepaspectratio]{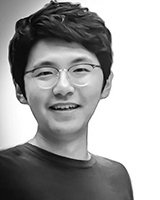}}]{Kyunghyun Paeng} received BS degree in 2011 in School of Electrical Engineering, KAIST, South Korea. Currently, he is a PhD student in the same school and department, and also a co-founder and a research scientist at Lunit Inc, South Korea. His research interests include 3D computer vision, visual recognition, weakly supervised learning, and medical image analysis with deep learning.
\end{IEEEbiography}
\begin{IEEEbiography}[{\includegraphics[width=1in,height=1.25in,clip,keepaspectratio]{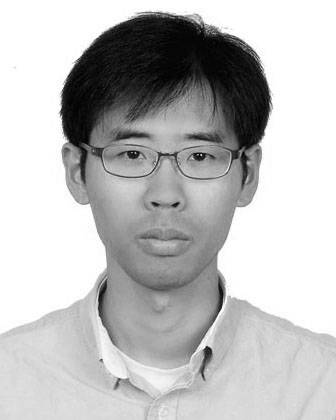}}]{Joon-Young Lee} received BS degree in electrical and electronic engineering from Yonsei University, South Korea, in 2008, and MS degree and PhD degree in 2009 and 2015 respectively, in School of Electrical Engineering from KAIST, South Korea. He is currently working in Adobe Research, San Jose, CA. His research interests include photometric methods in computer vision, image enhancement, computational photography, and deep learning for video analysis. He received the Samsung HumanTech Paper Award and the Qualcomm Innovation Award. He is a member of the IEEE.
\end{IEEEbiography}
\begin{IEEEbiography}[{\includegraphics[width=1in,height=1.25in,clip,keepaspectratio]{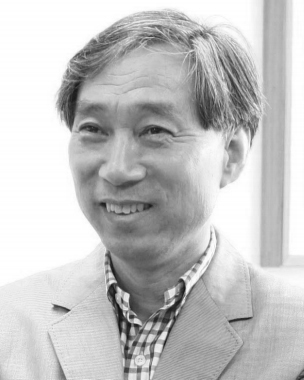}}]{In So Kweon} received BS and MS degrees in mechanical design and production engineering from Seoul National University, Korea, in 1981 and 1983, respectively, and PhD degree in robotics from the Robotics Institute, Carnegie Mellon University, Pittsburgh, PA, 1990. He was with Toshiba R\&D Center, Japan, and joined the Department of Automation and Design Engineering, KAIST, Korea in 1992, where he is currently a professor with the Department of Electrical Engineering. He received the best student paper runner-up award at the IEEE Conference on Computer Vision and Pattern Recognition (CVPR 09). His research interests are in camera and 3-D sensor fusion, color modeling and analysis, visual tracking, and visual SLAM. He was the program co-chair for the Asian Conference on Computer Vision (ACCV 07) and was the general chair for the ACCV 12. He is also on the editorial board of the International Journal of Computer Vision. He is a member of the IEEE and the KROS.
\end{IEEEbiography}





\end{document}